\documentclass[10pt,twocolumn,letterpaper]{article}

\usepackage[pagenumbers]{iccv} % To force page numbers, e.g. for an arXiv version
\usepackage[accsupp]{axessibility}

\newcommand{\method}{MultiverSeg\xspace}

\newcommand{\subpara}[1]{\vspace{0.4em} \noindent \textbf{#1.}}

\usepackage{algorithm}
\usepackage{algpseudocode}
\usepackage{booktabs}
\usepackage{amsmath}
\usepackage{bbm}
\usepackage{multirow}
\usepackage{colortbl}

\newcommand{\firstone}[1]{\colorbox{red!15}{#1}}
\newcommand{\secondone}[1]{\colorbox{blue!15}{#1}}

\makeatletter
\newif\ifcvpr@inappendix
\cvpr@inappendixfalse

\newcommand{\CVPRAppendixTOCMarker}{}

\newcommand{\MarkAppendixInTOC}{%
  \addtocontents{toc}{\protect\CVPRAppendixTOCMarker}%
}

\newcommand{\AppendixOnlyTOC}{%
  \begingroup
    \let\cvpr@orig@section\l@section
    \let\cvpr@orig@subsection\l@subsection
    \let\cvpr@orig@subsubsection\l@subsubsection
    \let\cvpr@orig@paragraph\l@paragraph
    \let\cvpr@orig@subparagraph\l@subparagraph
    \renewcommand{\l@section}[2]{\ifcvpr@inappendix \cvpr@orig@section{##1}{##2}\fi}
    \renewcommand{\l@subsection}[2]{\ifcvpr@inappendix \cvpr@orig@subsection{##1}{##2}\fi}
    \renewcommand{\l@subsubsection}[2]{\ifcvpr@inappendix \cvpr@orig@subsubsection{##1}{##2}\fi}
    \renewcommand{\l@paragraph}[2]{\ifcvpr@inappendix \cvpr@orig@paragraph{##1}{##2}\fi}
    \renewcommand{\l@subparagraph}[2]{\ifcvpr@inappendix \cvpr@orig@subparagraph{##1}{##2}\fi}
    \renewcommand{\CVPRAppendixTOCMarker}{\global\cvpr@inappendixtrue}
    \tableofcontents
  \endgroup
}
\makeatother

\definecolor{iccvblue}{rgb}{0.21,0.49,0.74}
\usepackage[pagebackref,breaklinks,colorlinks,allcolors=iccvblue]{hyperref}

\def\paperID{6615} % *** Enter the Paper ID here
\def\confName{ICCV}
\def\confYear{2025}

\title{MultiverSeg: Scalable Interactive Segmentation of Biomedical Imaging Datasets with In-Context Guidance}

\author{Hallee E. Wong\\
MIT CSAIL \& MGH\\
{\tt\small hallee@mit.edu}
\and
Jose Javier Gonzalez Ortiz\\
Databricks\\
{\tt\small josejg@mit.edu}
\and
John Guttag\\
MIT CSAIL\\
{\tt\small guttag@mit.edu}
\and 
Adrian V. Dalca\\
MIT CSAIL \& MGH,HMS\\
{\tt\small adalca@mit.edu}
}

\begin{document}
\twocolumn[{
\renewcommand\twocolumn[1][]{#1}%
\maketitle
\centering
\includegraphics[width=0.81\linewidth]{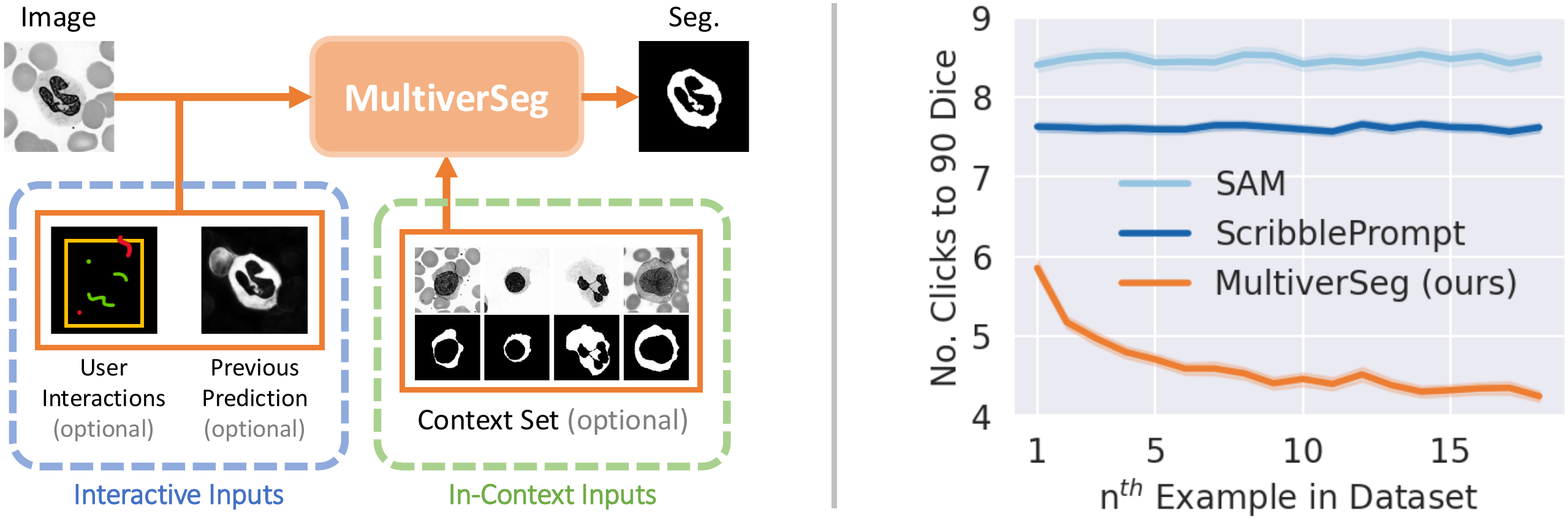}
\captionof{figure}{
\textbf{\method enables users to rapidly segment new datasets}. The \method network takes as input an image to segment, user interactions, and a context set of previously segmented image-segmentation pairs (\textbf{left}). As the user completes more segmentations, those images and segmentations become additional inputs to the model, populating the context set. As the context set of labeled images grows, the number of interactions required to achieve an accurate segmentation decreases (\textbf{right}). 
\vspace{2em}
}
\label{fig:teaser}
}]
\begin{abstract}
    Medical researchers and clinicians often need to perform novel segmentation tasks on a set of related images. Existing methods for segmenting a new dataset are either interactive, requiring substantial human effort for each image, or require an existing set of previously labeled images.

    We introduce a system, \textbf{MultiverSeg}, that enables practitioners to rapidly segment an entire new dataset without requiring access to any existing labeled data from that task or domain. Along with the image to segment, the model takes user interactions such as clicks, bounding boxes or scribbles as input, and predicts a segmentation. As the user segments more images, those images and segmentations become additional inputs to the model, providing context. As the context set of labeled images grows, the number of interactions required to segment each new image decreases. 
    
    We demonstrate that MultiverSeg enables users to interactively segment new datasets efficiently, by amortizing the number of interactions per image to achieve an accurate segmentation. 
    Compared to using a state-of-the-art interactive segmentation method, MultiverSeg reduced the total number of clicks by 36\% and scribble steps by 25\% to achieve 90\% Dice on sets of images from unseen tasks. 
    We release code and model weights at \href{https://multiverseg.csail.mit.edu}{https://multiverseg.csail.mit.edu}.
    \vspace{-\baselineskip}
\end{abstract}

%--------------------------------------------------------------------------------------
\section{Introduction}
%--------------------------------------------------------------------------------------

Segmentation is an important step in biomedical image analysis pipelines. Biomedical and clinical researchers often acquire novel images types or identify new regions of interest, and need to perform new segmentation tasks. Typically, scientists want to segment the same region of interest in many similar images from a new dataset.

Manually segmenting images is labor-intensive and requires domain expertise. \emph{Interactive} segmentation systems, in which a user provides a few clicks or scribbles on an image to produce a predicted segmentation, help to speed up the annotation of individual images. But with existing interactive segmentation systems, the user must independently repeat the same process for each image~\cite{SAM, wong2023scribbleprompt, MedSAM, MIDeepSeg, cheng_sam-med2d_2023, ITK-SNAP}. Ideally, a system should be able to learn from experience, becoming more accurate as the user completes more segmentations from the same task.

We propose \emph{\method}, a new interactive system that, as more images are segmented, progressively reduces the number of user interactions needed to predict accurate segmentations~(\cref{fig:teaser}). \method takes as input user interactions for a new image, along with a \textit{context set}, of example (previously segmented) image-segmentation pairs. 
To segment a new dataset, the user begins by interactively segmenting the first image. Once completed, the example becomes an input to \method, providing context for the segmentation of subsequent examples. The user interactively segments the next image with \method using bounding boxes, clicks or scribbles. As the user labels more images, the context set grows and the number of interactions required to achieve the desired segmentation of subsequent images decreases, often to zero. Unlike existing interactive segmentations systems~\cite{wong2023scribbleprompt, SAM, MIDeepSeg, MedSAM, zhang_iog_2020, DeepIGeoS, simpleclick, liu2024rethinking, xu_deep_2017, sakinis_IF-Seg_2019, wang_bifseg_2018} where the work required to segment a dataset is linear in the number of images, \method enables users to rapidly segment entire datasets. 

\noindent This paper
\begin{itemize}
    \item Presents \method, a new interactive segmentation framework that progressively reduces the amount of user interaction needed to predict accurate segmentations, as more images are segmented in a particular task.
    \item Introduces a model that segments an image given user prompts (bounding boxes, clicks and/or scribbles) and a variably-sized context set of previously labeled example images and segmentations. This network enables scalable segmentation of datasets by performing \emph{interactive} segmentation \emph{in context}.
    \item Demonstrates that \method can dramatically reduce the total number of user interactions needed to segment a collection of medical images.
\end{itemize}

%--------------------------------------------------------------------------------------
\section{Related Work}
%--------------------------------------------------------------------------------------

\subpara{Interactive Segmentation}
Recent interactive segmentation models can generalize to new segmentation tasks in medical~\cite{wong2023scribbleprompt, MedSAM, sammed2d-data} and natural images~\cite{SAM, SAM2}. While these methods are effective for segmenting single images, they require prohibitively extensive human interaction when segmenting large datasets. To incorporate new information, they must be retrained or fine-tuned. In contrast, at inference time \method can be conditioned on example segmentations from the new task, dramatically reducing the human effort required for accurate segmentation. 

Many works have sought to improve task-specific interactive segmentation performance by fine-tuning foundation models, either through full fine-tuning~\cite{kim_evaluation_2023} or more efficient adaption techniques~\cite{medsamadapter, paranjape2024adaptivesam, lin_samus_2023, wang_samocta_2023, hu_efficiently_2023, zhang_customized_2023, unclesam}. The fine-tuning must be repeated for each new task or group of tasks. It requires many relevant annotated images and the substantial computational resources needed to train a large model.

\subpara{In-Context Learning}
Recent \emph{in-context learning} approaches to segmentation~\cite{universeg, rakic2024tyche, wang2023seggpt, wu2024one, czolbe2023neuralizer, iclsam} enable users to perform new tasks by providing a set of labeled examples to the model at inference time. 
These methods often need existing large context sets to achieve adequate performance, and provide no mechanism for correcting predicted segmentations. 

A few works have explored in-context segmentation using a context set of example images with user annotations on those example images.
OnePrompt~\cite{wu2024one} segments a medical image given exactly one context example with click, scribble, bounding box or mask annotation. 
In contrast, \method has mechanisms to facilitate a variable number of context set entries and enables users to incorporate interactions on the target image along with the context of previous segmentations, enabling substantially richer use cases.
LabelAnything~\cite{LabelAnything} is a few-shot framework that enables multi-label segmentation of a target natural image given a small context set of example images with click, box, or mask annotations on the examples. 
In contrast, \method can leverage \textit{large} context sets, and enables users to provide interactions on the target image and corrections to refine the prediction to get the desired result.

\subpara{Continual Learning}
One approach to segmenting a new dataset involves manually labeling a large number of images, and then training an automatic task-specific model~\cite{isensee_nnunet_2021} to segment the rest. 
For example, \mbox{MonaiLabel}~\cite{monailabel} is an open-source tool that packages this process. In contrast, \method can be adapted \textit{at inference time} using new labels (collected manually or interactively) without the need to re-train.

\subpara{Annotation-Efficient Learning}
Another approach to segmenting a new dataset is to collect sparse annotations on many images and train an automatic segmentation model using these annotations for supervision. The annotations can be bounding boxes~\cite{lan2021discobox, tian2021boxinst}, clicks~\cite{liu2023clickseg, liu2021onethingoneclick, roth_going_2021} or scribbles~\cite{lin2016scribblesup, li_scribblevc_2023, zhang2022cyclemix, luo_scribble_2022, gotkowski2024embarrassingly}. These methods require the manual annotation of many training images and retraining for each new task. Other methods use annotations to perform online learning, using user corrections as a source of supervision to update the model weights at test time~\cite{sofiiuk_fbrs_2020, asad_adaptive_2023, asad_econet_2022, wang_dynamically_2016, kontogianni2020continuous, zheng2021continual}.
In contrast, \method is trained only once on a large corpus of datasets, and then can be used to segment new datasets at inference time without any retraining, using graphical interactions and previous segmentations as \emph{input}.

%--------------------------------------------------------------------------------------
\section{Methods}
%--------------------------------------------------------------------------------------

\begin{figure*}[ht]
    \centering
    \includegraphics[width=\linewidth]{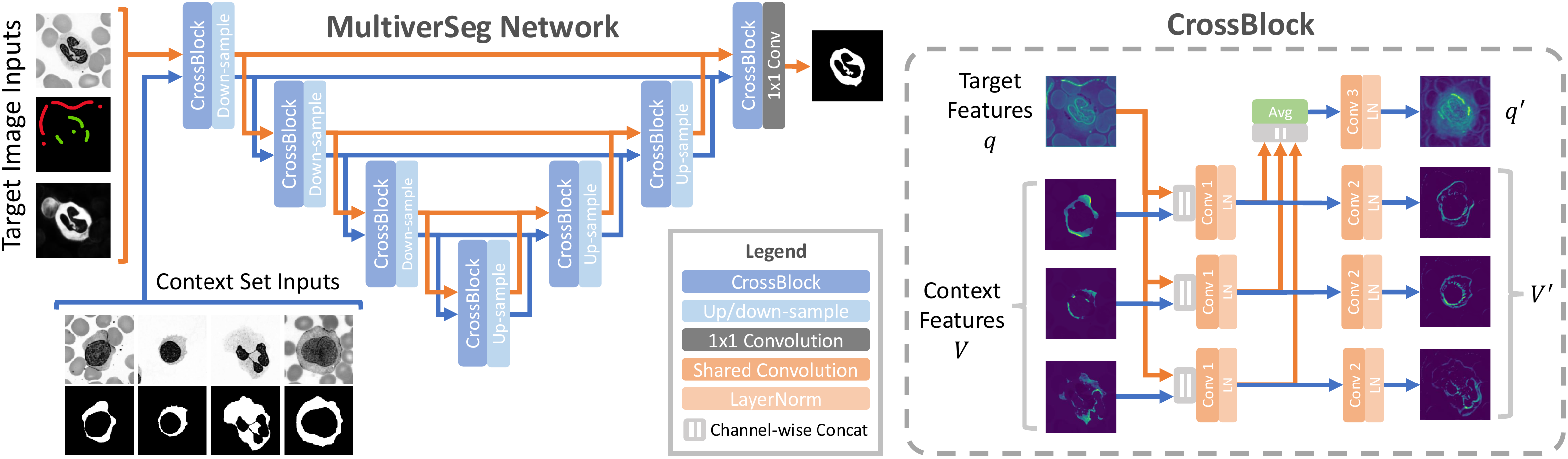}
    \caption{\textbf{\method Architecture}. The \method network (\textbf{left}) takes as input a stack of target image inputs $q_i$ and a context set of image-segmentation pairs $\{(x_l, y_l)\}_{l=1}^m$. The target image inputs include a target image $x_i$, optional user interactions $u_{i,j}$, and a previous predicted segmentation $\hat{y}_{i,j-1}$, if available. The architecture is similar to a UNet~\cite{unet}. However, we use a CrossBlock~\cite{universeg} (\textbf{right}) with additional normalization layers~\cite{LayerNorm} to interact the features of the target image inputs $q_i$ with the features of the context set inputs $V = \{v_l\}_{l=1}^m$ throughout the network.
    }
    \label{fig:architecture}
\end{figure*}

\subsection{Problem Setup}

For a new task~$t$, we aim to segment a set of images $\{x_i^t\}_{i=1}^N$ into their corresponding segmentations $\{y_i^t\}_{i=1}^N$. 

We assume a user provides manual interactions for one image at a time, to indicate the desired segmentation. These interactions may be iterative even for a single image, sometimes indicating corrections based on a previous prediction $\hat{y}_{i,j-1}^t$. We let $u_{i,j}^t$ be the interactions provided for image $x_i$ at step $j$, and $\hat{y}_{i,k_i}^t$ be the final predicted segmentation for image $x_i$ after $k_i$ steps of interaction.

We want to maximize the quality of predicted segmentations $\{\hat{y}_{i,k_i}^t\}_{i=1}^N$ for an entire dataset, $\min \sum_{i=1}^N \mathcal{L}_{seg}(y_i^t, \hat{y}_{i,k_i}^t)$ while minimizing the \textit{total} number of user interactions $\sum_{i=1}^N \sum_{j=1}^{k_i} u_{i,j}^t$, where $\mathcal{L}_{seg}$ is a segmentation distance metric, $\{y_i^t\}_{i=1}^N$ are the ground truth segmentation maps, and $k_i$ is the number of steps of interaction for image $i$.

\subsection{\method}

We introduce a framework that enables rapid progressive segmentation of an entire dataset. The key component is a method that segments image~$x^t_i$ based on user interactions $u_{i,j}$ and a set of image-segmentation maps $S_i^t = \{(x_l^t, \hat{y}^t_{l,k_l}) \}_{l=1}^{i-1}$ for previous segmented images. 

\subpara{First image} When segmenting the first image~$x_0^t$ of a new task~$t$, the context set~$S_0^t$ is empty. We build on learning-based interactive segmentation approaches~\cite{SAM, wong2023scribbleprompt}, to learn a function $g_\phi(x_i^t; u_i, \hat{y}^t_{i,j-1})$ that at step $j$ produces a segmentation $\hat{y}_{i,j}$ of image $x^t$, given a set of user interactions $u_{i,j}$ and a previously predicted segmentation $\hat{y}^t_{i,j-1}$. The interactions $u_i$, which may include positive or negative scribbles, positive or negative clicks, and bounding boxes, are provided by a user who has access to the image $x^t$ and previous prediction $\hat{y}_{i-1}^t$. For $g_\phi(\cdot)$, we use the pre-trained ScribblePrompt-UNet model~\cite{wong2023scribbleprompt}.

\subpara{Subsequent images} For subsequent images $x_{i>0}^t$, the context set $S_i^{t} = \{(x_l^{t}, \hat{y}_{l,k_l}^{t})\}_{l=0}^{i-1}$ encompass the previously segmented images and the resulting segmentation maps. We learn a function $f_\theta(x^t ; u_i, \hat{y}^t_{i-1}; S^t)$ with parameters $\theta$ that leverages the set of user interactions $u_i$, previous prediction $\hat{y}^t_{i-1}$, and context set~$S^t_i$ to produce a segmentation $\hat{y}_i$. As more images are segmented, the context set~$S_i^t$ grows, leading to fewer interactions~$u_i$ needed for accurate segmentation of each subsequent image.

\subsubsection{Architecture}

We employ a convolutional architecture (\cref{fig:architecture}) for $f_\theta$ with an encoder-decoder structure, building on recent in-context learning strategies~\cite{universeg}. The architecture uses a \textit{CrossBlock} mechanism to mix information between the context set, which can be of variable size, and the inputs corresponding to the user interactions, which pertain to the target image. 

\subpara{Target image inputs}
The target image inputs consist of the target image $x_i$ and graphical user interactions $u_{i,j}$, and a previous prediction $\hat{y}_{i,j-1}$, if available. The interactions may include bounding boxes, positive clicks and scribbles, and negative clicks and scribbles, represented as three intensity-based masks~\cite{wong2023scribbleprompt}. We stack the target inputs leading to five channels, where the first channel contains the input image, and the other channels contain the optional interactions and an optional previous prediction. When there are no interactions or there is no previous prediction, these channels are set to zero. 

\subpara{Context set inputs}
For each of the $N$ examples in the context set, we stack the image and segmentation.

\subpara{CrossBlock}
We use a modified CrossBlock mechanism to interact intermediate target features $q$ with intermediate features of the context set inputs $v$~\cite{universeg}. 
 
We use a \emph{cross-convolution} layer to interact a target feature map $q$ with a set of context feature maps $V = \{v_i \}_{i=1}^n$:
\begin{equation}
\begin{aligned}
    &\text{CrossConv}(q, V; \theta_z) = \{z_i\}_{i=1}^n,\\
    &\qquad\text{for}\; z_i = \text{Conv}(q || v_i; \theta_{z}) .
\end{aligned}
\end{equation}
This layer is used within the Crossblock to produce features of target representation $q$ and context set $V$ at each step in the network:
\begin{align}\label{eq:cross_block}
    &\text{CrossBlock}(q, V; \ \theta_{z,q,v}) = (q^\prime, V^\prime), \text{where:} \\
    &\quad z_i = \text{LN}(A(\text{CrossConv}(q, v_i; \theta_z))) ~ \text{for}\; i = 1,2,\ldots ,n \nonumber
    \\
    &\quad q^\prime = \text{LN}(A(\text{Conv}(1/n \textstyle\sum_{i=1}^n z_i; \theta_q))) \nonumber
    \\
    &\quad v^\prime_i = \text{LN}(A(\text{Conv}(z_i; \theta_v))) \quad \text{for}\; i = 1,2,\ldots ,n, \nonumber 
\end{align}
where $A(x)$ is a non-linear activation function and $LN(\cdot)$ is LayerNorm~\cite{LayerNorm}. 

\subpara{Network} We employ a UNet-like encoder-decoder architecture, where each convolutional block is replaced by a CrossBlock, enabling the target-related inputs to interact with the previously segmented images at every image scale~\cite{unet, universeg}. 

\subsection{Training}

We summarize the training process in \cref{alg:training-loop-independent}. During training, we first sample a random task~$t$, and then sample a training example $(x_i^t, y_i^t)$ and context set $S_i^t = \{(x_l^t, y_l^t)\}_{l=0}^n$ of random size $n \in [0,N]$. We use ground truth segmentation labels in the context set during training. 

We minimize the difference between the true segmentation $y^t$ and each of the $k$ iterative predictions $\hat{y}_{i,0}, \dots, \hat{y}_{i,k}$, given a context set $S^t$, 
\begin{equation}\label{eq:loss}
    \small
    \mathcal{L}(\theta; \mathcal{T}) = 
    \mathbb{E}_{t \in \mathcal{T}}
    \left[ \mathbb{E}_{(x_i^t,y_i^t; S^t) \in t} 
    \left[ \sum_{j=1}^k \mathcal{L}_{seg} \left(y^t_i, \hat{y}^t_{i,j} \right) \right] \right],
\end{equation}
where $\hat{y}^t_{i,j} = f_\theta(x^t_i, u^t_{i,j}, \hat{y}^t_{i,j-1}; S_i^t)$,  $\mathcal{L}_{seg}$ is a supervised segmentation loss, $x_t \notin S^t$ and $\hat{y}^t_{0} = \mathbf{0}$. 

\begin{algorithm}[t]
\caption{\method Training Loop using SGD with learning rate~$\eta$ over tasks~$\mathcal{T}$ with independently sampled context set, main architecture~$f_\theta$, in-task augmentations~$\text{Aug}_t$ and task augmentations~$\text{Aug}_T$}
\label{alg:training-loop-independent}
\begin{algorithmic}
\For{$k = 1,\ldots, \text{NumTrainSteps}$}
\State $t\sim \mathcal{T}$ \Comment{Sample Task}
\State $(x_i^{t}, y_i^{t}) \sim t$ \Comment{Sample Target}
\State $n \sim U[0,N]$ \Comment{Sample Context Size}
\State $S^{t} \gets \{ (x_l^{t}, y_l^{t})\}_{l\neq i}^n$ \Comment{Sample Context}
\State $x_i^{t}, y_i^{t} \gets \text{Aug}_t(x_i^{t}, y_i^{t})$ \Comment{Augment Target}
\State $S^{t} \gets \{ \text{Aug}_t(x_l^{t}, y_l^{t}) \}_l^n$ \Comment{Augment Context}
\State $x_i^{t}, y_i^{t}, S^{t}\gets \text{Aug}_T(x_i^t, y_i^{t}, S^{t})$ \Comment{Task Aug}
\State $\hat{y}_{i,0} \gets \mathbf{0}$ 
\For{$j = 1, \ldots, \text{NumInteractionSteps}$}
\State $u_{i,j}^t \gets h_\psi(y_i^t, \hat{y}_{j-1})$ \Comment{Simulate Interactions}
\State $\hat{y}_{i,j} \gets f_\theta(x_i^{t}, u_{i,j}^t, \hat{y}_{i,j-1}; S^{t})$ \Comment{Predict Seg.}
\State $\ell_j \gets \mathcal{L}_{\text{seg}}({y}_i^{t},\hat{y}_{i,j})$ \Comment{Compute Loss}
\EndFor
\State $\theta \gets \theta - \eta \nabla_\theta \sum_j \ell$ \Comment{Gradient Step}
\EndFor
\end{algorithmic}
\end{algorithm}

\subpara{Prompt Simulation}
We simulate random combinations of scribbles, clicks and bounding boxes during training following the prompt simulation procedures described in \cite{wong2023scribbleprompt}. We simulate $k$ steps of interactive segmentation for each example during training. For the first step ($i=1$), we sample the combination of interactions (bounding box, clicks, scribbles) and the number of initial positive and negative interactions $n_{pos},n_{neg} \sim U[n_{min},n_{max}]$. The initial interactions $u_1$ are simulated using the ground truth label $y^t$. In subsequent steps, we sample correction scribbles or clicks from the error region $\varepsilon_{i-1}^t$ between the last prediction $\hat{y}^t_{i-1}$ and the ground truth $y^t$. Since a user can make multiple corrections in each step, we sample $n_{cor} \sim U[n_{min},n_{max}]$ corrections (scribbles or clicks) per step. 

%--------------------------------------------------------------------------------------
\section{Data}
%--------------------------------------------------------------------------------------

\subpara{Task Diversity}
We use a collection of 79 biomedical imaging datasets (\cref{appendix:data}) and synthetically generated images and tasks~\cite{universeg,rakic2024tyche,wong2023scribbleprompt}. 
The collection includes a diverse array of biomedical domains, such as eyes~\cite{STARE, OCTA500, Rose, IDRID, DRIVE}, thorax~\cite{CheXplanation, LUNA, MSD, PAXRay, SCD}, spine~\cite{SpineWeb, VerSe, PAXRay}, cells~\cite{WBC, BBBC003, BBBC038, ssTEM, ISBI_EM, PanNuke, CoNSeP}, skin~\cite{ISIC}, abdomen~\cite{LiTS, NCI-ISBI, KiTS, AMOS, CHAOS_1, SegTHOR, BTCV, I2CVB, Promise12, Word, SCD, MSD, DukeLiver, CT2US, CT_ORG}, neck~\cite{SegThy, HaN-Seg, NerveUS, DDTI}, brain~\cite{BRATS, MCIC, ISLES, WMH, BrainDevelopment, OASIS-data, PPMI, LGGFlair, MSD, COBRE}, bones~\cite{PAXRay, HipXRay, SCR, TotalSegmentator}, teeth~\cite{PanDental, ToothSeg} and lesions \cite{BUID, BUSIS, MMOTU, MSD}.

\subpara{Task Definition}
We define a 2D segmentation task as a combination of dataset, modality, axis (for 3D modalities), and binary label. For datasets with multiple segmentation labels, we consider each label as a binary segmentation task and for 3D modalities we use the slice with maximum label area and the middle slice from each volume.   

\subpara{Data Augmentation} 
We perform both task augmentation and within-task data augmentation to increase the diversity of segmentation tasks~\cite{universeg}. For task augmentation, the same augmentation is applied to the target example and the entries of the context set to change the segmentation task. For within-task augmentation, we apply data augmentation where the parameters are randomly sampled for each target example and context set entry, to vary the examples within a task. Augmentations are applied prior to simulating the user interactions. We detail the augmentations in \cref{appendix:data_aug}.

\subpara{Synthetic Data}
Synthetic data can help improve generalization~\cite{universeg, wong2023scribbleprompt, gopinath2024synth, billot2023synthseg}. We use fully synthetic data (images and labels) similar to strategies used for in-context learning~\cite{universeg}. 

\subpara{Synthetic Tasks}
We introduce a new approach for constructing synthetic tasks from real images. Given a single image $x_0$ we construct a set of images $\{x_i', y_i'\}_{i=1}^{m+1}$ representing a synthetic task. We then partition this set into a target example and context set of size $m$ for training. 

Given an image $x_0$, we first generate a synthetic label $y_{synth}$ by applying a superpixel algorithm \cite{felzenszwalb_superpixels} with scale parameter $\lambda \sim U[1,\lambda_{max}]$ to partition the image into a multi-label mask of $k$ superpixels $z \in \{1, \dots, k\}^{n \times n}$. We then randomly select a superpixel $y_{synth} = \mathbbm{1} (z = c)$ as a synthetic label. 

To generate a set of $m+1$ images representing the same task, we duplicate $(x_0, y_{synth})$, $m+1$ times and apply aggressive augmentations to vary the images and segmentation labels~\cite{zhao2019dataaug,universeg}. We detail these augmentations and provide examples in \cref{appendix:synth_aug}. 

During training, we replace a randomly sampled target example $(x_0^t, y_0^t)$ and context set $S^t$ with synthetic ones with probability $p_{synth}$. We use $p_{synth}=0.5$.

%--------------------------------------------------------------------------------------
\section{Experimental Setup}
%--------------------------------------------------------------------------------------

We evaluate \method and baselines in segmenting a set of images, representing a segmentation task unseen during training. We simulate the process of interactively segmenting each image in a dataset, and of adding the segmentations to the context set as they are completed.

\subsection{Training \method}

To learn $f_\theta(\cdot)$, we minimize eq. (\ref{eq:loss}) where $\mathcal{L}_{seg}$ is the sum of soft Dice Loss \cite{dice1945measures} and Focal Loss \cite{focal_loss} with $\gamma=20$~\cite{SAM}. We minimize the loss using the Adam optimizer~\cite{kingma2014adam}. 
We simulate 3 steps of interactive segmentation for each example during training. We simulate 1-3 positive and 0-3 negative interactions in the first step, and 1-3 corrections per subsequent step.
We randomly sample a context set of size $m \sim U[0,64]$ for each sample, and train with a batch size of $2$ and learning rate of $\eta=10^{-4}$.  

\subsection{Data}

We partition our collection of 79 datasets into 67 datasets for training and 12 datasets held-out for evaluation. 
We report results on the 12 held-out datasets that were unseen by the model during training. These datasets cover 187 tasks and 8 modalities, including unseen image types, anatomies, and labels. The evaluation datasets cover a variety of modalities (MRI, CT, ultrasound, fundus photography, microscopy) and anatomical regions of interest (brain, teeth, bones, abdominal organs, muscles, heart, thorax, cells), including both healthy anatomy and lesions~\cite{ACDC,BTCV,SCR,SCD,BTCV,BUID,STARE,SpineWeb,HipXRay,WBC,COBRE,TotalSegmentator}. 

\subsection{Prompt Simulation}

Throughout our experiments, we consider two inference-time interaction protocols:
\begin{itemize}
    \item \textbf{Center Clicks}: One positive click in the center of the largest component to start (step 1), followed by one (positive or negative) correction click per step in the center of the largest component of the error region. 
    \item \textbf{Centerline Scribbles}: One positive and one negative centerline scribble to start (step 1), followed by one positive or negative correction centerline scribble per step. 
\end{itemize}

\noindent We selected these protocols because center clicks are commonly used for evaluation~\cite{RITM, xu_dios_2016, simpleclick, huang2023interformer, liu2024rethinking, SAM, benenson_largescale_2019} and centerline scribbles were the most effective prompt in~\cite{wong2023scribbleprompt}.

\subsection{Baselines}

\subpara{Interactive Segmentation Baselines}
We compare to five interactive segmentation methods trained on biomedical images: ScribblePrompt~\cite{wong2023scribbleprompt}, MedSAM~\cite{MedSAM}, SAM-Med2D~\cite{cheng_sam-med2d_2023}, and IMIS-Net~\cite{imis}. 
We also evaluated two general interactive segmentation methods, SAM~\cite{SAM} and SegNext~\cite{liu2024rethinking}, which were trained on natural images. 
We focus on ScribblePrompt over SAM, and medical imaging variants of SAM~\cite{MedSAM, cheng_sam-med2d_2023, imis}, because it produced more accurate segmentations on unseen biomedical imaging datasets~\cite{wong2023scribbleprompt,rokuss2025lesionlocator} and has faster inference runtime.

\subpara{In-Context Segmentation Baselines}
We compare to UniverSeg~\cite{universeg}, a general state-of-the-art in-context segmentation model that was trained on a diverse collection of biomedical images. 
We did not compare to OnePrompt~\cite{wu2024one} and LabelAnything~\cite{LabelAnything} because the pre-trained weights were not publicly available. We discuss further in \cref{appendix:baselines}.

\subpara{Interactive In-Context Segmentation Baselines}
We construct a new baseline, \emph{SP+UVS}, by combining UniverSeg~\cite{universeg} and ScribblePrompt~\cite{wong2023scribbleprompt}. We use the publicly available pre-trained weights for each model. When the context set is empty, we use ScribblePrompt. When provided with a context set, we first predict using UniverSeg, and then refine the prediction with ScribblePrompt.

Consistent with the original published results, we find that UniverSeg has poor performance for small context sets and initializing ScribblePrompt using the UniverSeg prediction hurts performance when the context set is small (\cref{fig:val_dataset_per_data_variations}). Thus, for context sets with fewer than 5 examples, we ignore the context and use only ScribblePrompt to make predictions.

\subpara{Supervised Benchmarks (upper bound)}
We also train task-specific models using the popular nnUNet pipeline~\cite{isensee_nnunet_2021}, which automatically configures the model architecture and training based on the data properties. We train a separate nnUNet model for each held-out 2D task, and report results from the collection of models. These models act as upper bounds on segmentation accuracy, because they are fully-supervised and have access to ground truth training data not available to the other algorithms. 

\subsection{Metrics}

We evaluate segmentation quality using Dice score~\cite{dice1945measures}, and show 95th percentile Hausdorff distance~\cite{huttenlocher_hd_1993} in \cref{appendix:interactions_per_example}.
\method, UniverSeg, and ScribblePrompt were all trained and developed on images at $128^2$ resolution. Unless otherwise noted, we evaluated on images resized to $256^2$ resolution to demonstrate performance at a higher, more realistic, resolution. 
We show results with similar trends in \cref{appendix:res128}, evaluating at $128^2$ resolution.

\begin{figure}
    \centering
    \includegraphics[width=\linewidth]{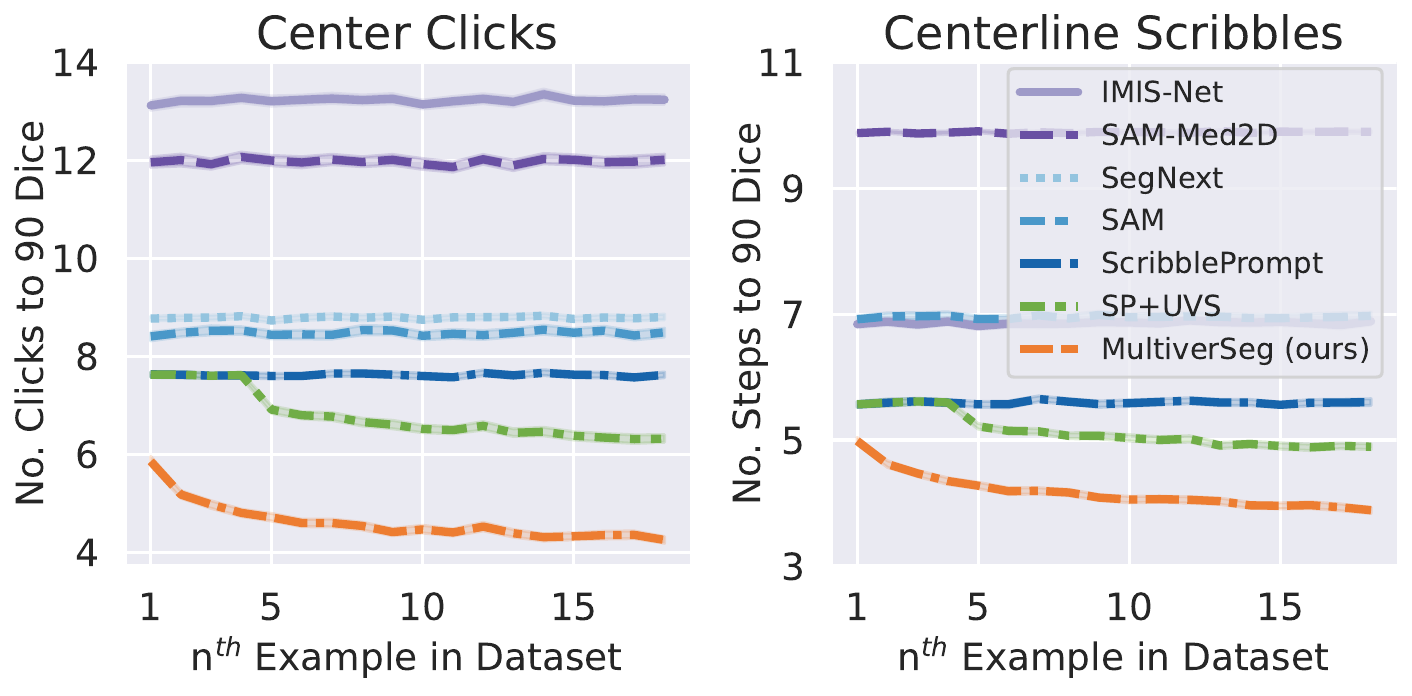}
    \caption{\textbf{Interactions to target Dice on unseen tasks}. Number of interactions needed to reach a 90\% Dice as a function of the example number being segmented. For the $n^{th}$ image being segmented, the context set has $n$ examples. \method requires substantially fewer number of interactions to achieve 90\% Dice than the baselines, and as more images are segmented, the average number of interactions required decreases dramatically.
    }
    \label{fig:dataset_interactions_vs_subject}
\end{figure}

\begin{figure}
    \centering
    \includegraphics[width=\linewidth]{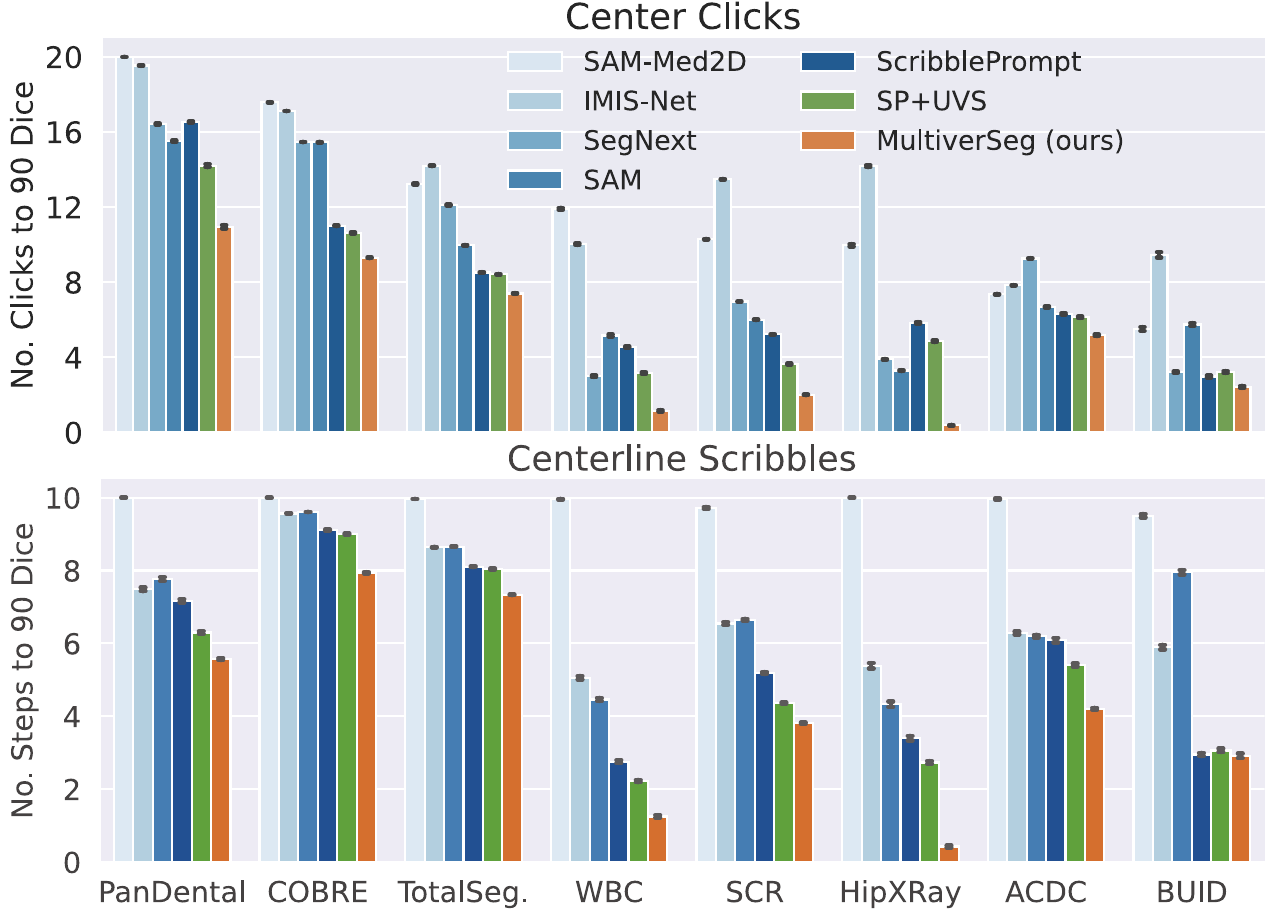}
    \caption{\textbf{Interactions per image by unseen dataset}. We show average number of clicks and scribble steps per image to segment 18 images to $\geq90\%$ Dice for each method. In all scenarios, \method required fewer or the same number of interactions than the best baseline. Error bars show 95\% CI from bootstrapping.}
    \label{fig:total-interactions}
\end{figure}

\begin{figure*}
    \centering
    \includegraphics[width=\textwidth]{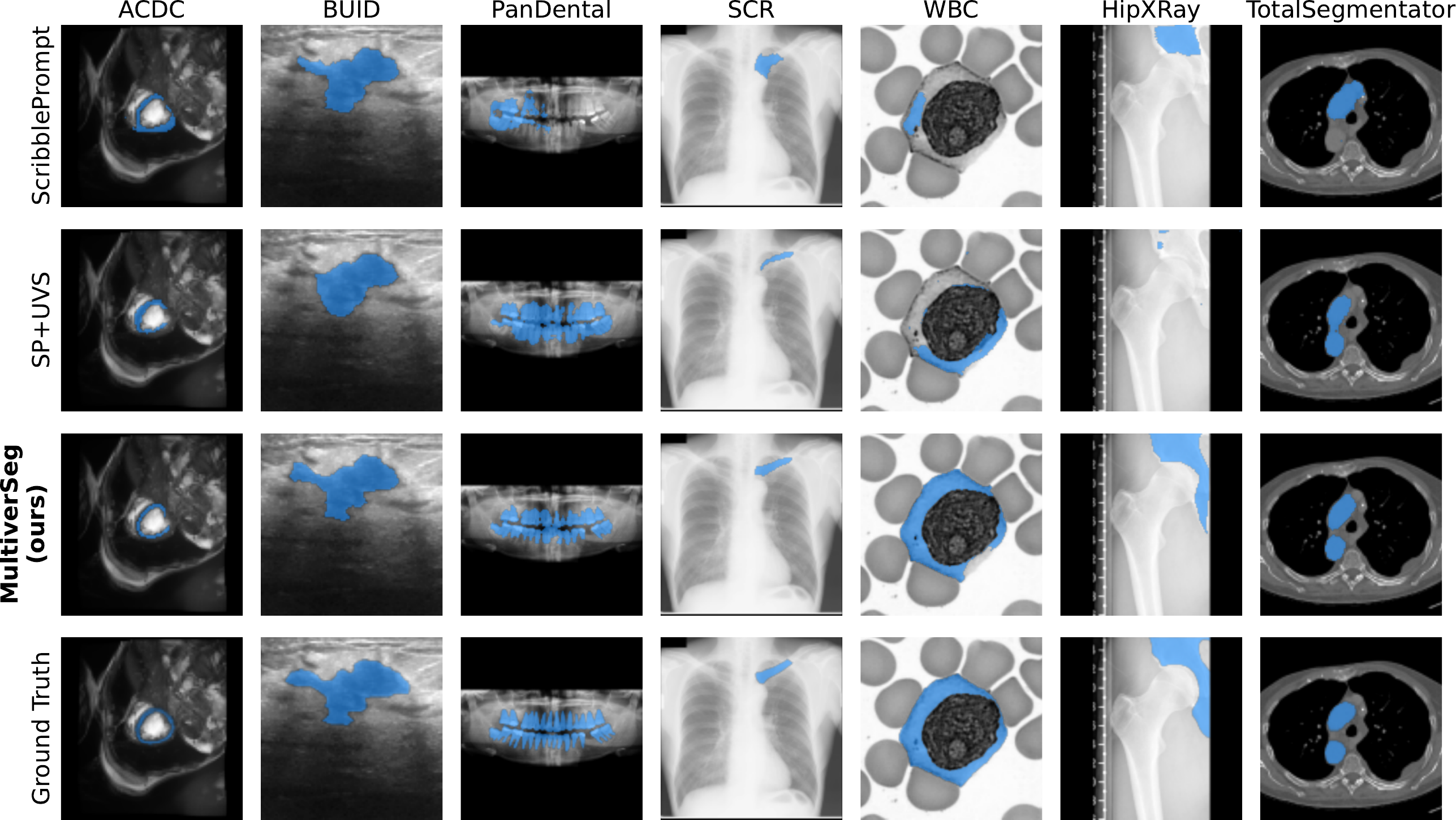}
    \caption{\textbf{Example predictions after 1 interaction step}. We show predictions for \method and the top two performing baselines on a randomly chosen example from each held-out task. We use a context set of 10 examples that were previously segmented to $\geq$90\% Dice. For each method, we show the prediction after 1 step of interaction: 1 step of centerline scribbles for ACDC~\cite{ACDC}, BUID~\cite{BUID}, and PanDental~\cite{PanDental}, and 1 center click for SCR~\cite{SCR}, WBC~\cite{WBC}, HipXRay~\cite{HipXRay}, and TotalSegmentator~\cite{TotalSegmentator}. 
    }
    \label{fig:qualitative_examples}
\end{figure*}

%--------------------------------------------------------------------------------------
\section{Experiment 1: Evaluation}
\label{sec:main_results}
%--------------------------------------------------------------------------------------

In this experiment, we evaluate different approaches to segmenting an entire new biomedical dataset. We compare \method to ScribblePrompt, SAM, SegNext, SAM-Med2D, IMIS-Net, and MedSAM, which perform interactive segmentation of each image independently, and to SP+UVS, which combines ScribblePrompt with an in-context segmentation model (UniverSeg). We show that \method outperforms all of the baselines. 

\subsection{Setup}

We evaluate the number of interactions required to achieve a target Dice score on each image using different methods, or a maximum number of interactions if the score was not reached. We use 90\% as a target Dice score, because our collection of fully-supervised task-specific nnUNet models achieves an average Dice of $88.67 \pm 0.47$ on the same test data. We set 20 center clicks or 10 steps of centerline scribbles as the maximum number of interactions.  

For each method and task, we begin by interactively segmenting one randomly sampled image from the training split to $\geq$ 90\% Dice using ScribblePrompt. This example is used to seed the context set. We then randomly sample (without replacement) 18 images from the test split, and simulate sequentially segmenting each image. 

\subpara{Data} 
We report results averaged across 200 simulations for each held-out segmentation task. We exclude tasks with fewer than 18 test examples, leaving 161 tasks from 8 evaluation datasets~\cite{COBRE,ACDC,BUID,PanDental,HipXRay,SCR,WBC,TotalSegmentator}. We further discuss the choice of this cutoff in \cref{appendix:experiment1_setup}. 

\subsection{Results}

\subpara{Interactions per image as a function of dataset size}
As more examples are segmented and the context set grows, the number of interactions required to get to $90\%$ Dice (NoI90) on the $\text{n}^\text{th}$ example using \method decreases substantially (\cref{fig:dataset_interactions_vs_subject}). For interactive segmentation methods, NoI90 is approximately constant, because they are not designed to learn from previous examples. With SP+UVS, the number of interactions decreases as more examples are segmented, but it requires more interactions than \method. Results by task in \cref{appendix:interactions_per_example} show a similar trend.
\cref{fig:qualitative_examples} shows predictions for the $10^{th}$ example after 1 step of interaction. 

\subpara{Total interactions}
On average, using \method reduced the number of clicks required to segment each dataset by $(36.41 \pm 1.33)\%$ and the number of scribbles steps required by $(25.26 \pm 1.80)\%$ compared to ScribblePrompt (\cref{fig:total-interactions}).
For larger sets of images, using \method results in even greater reductions in the total number of user interactions (\cref{appendix:interactions_per_example}).

\subpara{Other Baselines} MedSAM, which only works with bounding boxes, had an average Dice of $65.93 \pm 4.82$ and was only able to reach 90\% Dice for $5.6\%$ of examples. SegNext failed with scribbles due to GPU memory limits.

\subpara{Context Set Quality}
\method was trained with ground truth context set labels. However, at inference time, the context set only includes previously predicted segmentations. 
For both \method and SP+UVS, thresholding the predictions at 0.5 before adding them to the context set improved the accuracy of predictions for subsequent images. We show the effect of this modification in \cref{appendix:interactions_per_example}.

As an upper bound on performance, we also evaluated using ground truth labels in the context set instead of predicted segmentations (\cref{appendix:interactions_per_example}). Using ground truth context labels decreases the number of interactions to achieve $90\%$ Dice for both \method and SP+UVS, but \method still requires fewer interactions.

\subpara{Bootstrapping In-Context Segmentation}
Another approach to segmenting a new dataset is to manually segment an image, and then use an in-context segmentation model to segment the rest of the images.
We experimented with bootstrapping UniverSeg: starting from a single labeled example as the context set, we sequentially segment each image with UniverSeg and then add it to the context set for the next example. This approach did not produce accurate results ($48.89 \pm 1.87$ Dice), likely because UniverSeg has poor performance for small context sets (\cref{fig:eval_context_size}) and/or context sets with imperfect labels. Because UniverSeg does not have a mechanism to incorporate corrections, it was not possible to achieve 90\% Dice for most images. We show results experimenting with this approach in \cref{appendix:bootstrap_uvs}. 

\subpara{Task-Specific Fine-Tuning} Another approach is to interactively segment a few images, and then \textit{fine-tune} ScribblePrompt using those labeled examples to produce a \emph{task-specific} interactive segmentation model. This requires computational overhead and machine-learning expertise that is often unavailable in biomedical research or clinical workflows. As we show in \cref{appendix:finetuning}, even if it were practical, the fine-tuned models do not perform as well as \method. 
Fine-tuning each task-specific model took 20 minutes on average using a NVIDIA A100 GPU. In contrast, MultiverSeg's inference runtime is $<0.15$ seconds, even with a context set size of 64 examples (\cref{appendix:runtime}).

\subpara{Limitations}
MultiverSeg does not perform as well on tasks where the context set images vary substantially in composition, especially for limited set sizes. E.g., \cref{fig:clicks_to_dice_dataset} shows that with clicks, MultiverSeg underperforms ScribblePrompt on the BUID dataset until the context set has $\geq 5$ examples. Given scribbles, which provide more information, context is less helpful, and ScribblePrompt and MultiverSeg have similar performance on BUID (\cref{fig:scribbles_to_dice_dataset}).

%--------------------------------------------------------------------------------------
\section{Experiment 2: Analysis}
%--------------------------------------------------------------------------------------

When segmenting images sequentially, as in the previous experiment, the performance on the $n^{th}$ image is correlated with the predictions on the previous images. However, in some realistic instances, a few ground-truth segmentations might be available from other previous segmentation efforts. In the following experiments, we analyze \method using randomly sampled context sets with \textit{ground truth} labels. We report results on the test split of 12 evaluation datasets not used during training. 
Because of the computational burden of training 187 task-specific nnUNets, we trained on images with $128^2$ resolution. Thus, for this experiment, we report results at $128^2$ resolution.

\subsection{In-Context Segmentation}

\subpara{Setup}
We compare the predictions of \method to a generalizable in-context learning baseline, UniverSeg~\cite{universeg}, given different context set sizes. 
For each test example, we make 10 predictions with context sets randomly sampled with replacement from the training split of the same dataset. 

\begin{figure}
    \centering
    \includegraphics[width=\linewidth]{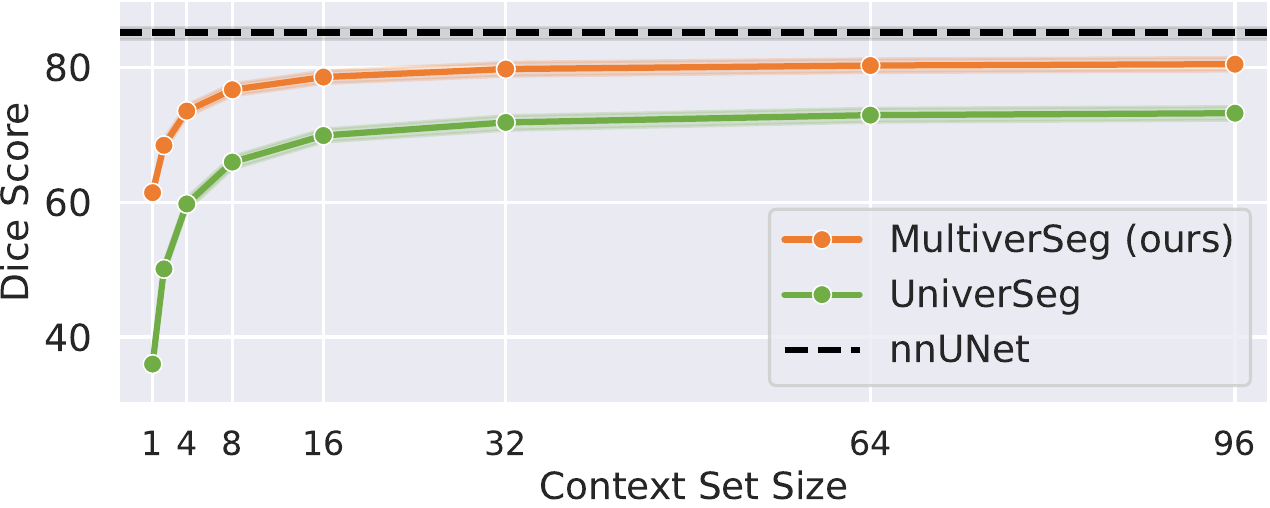}
    \caption{\textbf{In-context segmentation performance across context set sizes}. We compare \method to an in-context segmentation method, UniverSeg~\cite{universeg}, given ground truth context labels. Shading shows 95\% CI from bootstrapping.
    }
    \label{fig:eval_context_size}
\end{figure}

\subpara{Results}
\method produces higher Dice score segmentations than UniverSeg across all context set sizes (\cref{fig:eval_context_size}). In the previous experiment, \method required fewer interactions than SP+UVS, in part because its initial in-context predictions were more accurate than those of UniverSeg. This is likely due to \method being trained on a larger collection of data (67 vs. 53 datasets) with more features per CrossBlock (256 vs. 64) and normalization layers.

\begin{figure}
    \centering
    \includegraphics[width=\linewidth]{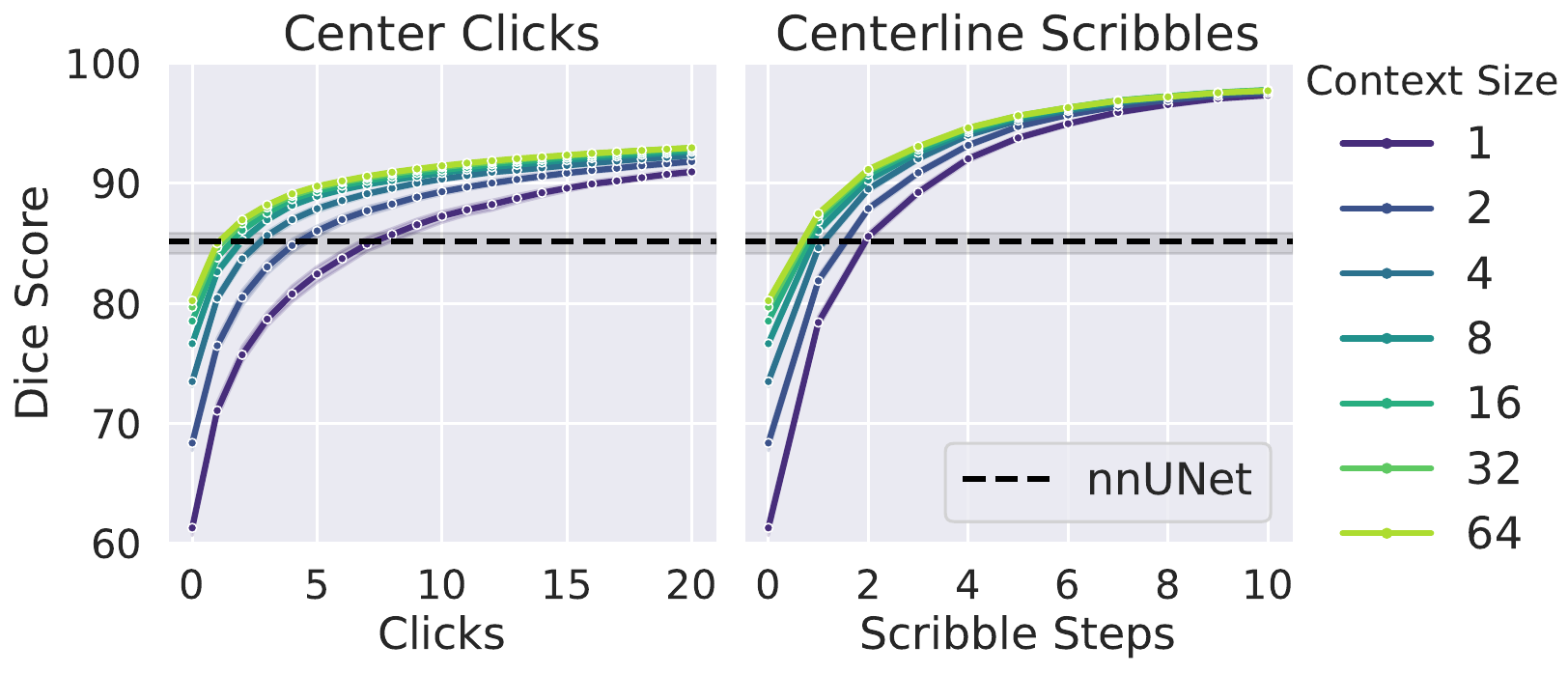}
    \caption{
    \textbf{Interactive segmentation in context}.
    \method's interactive segmentation performance improves as the context set size grows. We first make an initial prediction based on the context set (step 0), and then simulate corrections (clicks or scribbles). Shading shows 95\% CI from bootstrapping. 
    }
    \label{fig:interseg_context_size}
\end{figure}

\subsection{Interactive Segmentation in Context}

\subpara{Setup}
We evaluate the interactive segmentation performance of \method given context sets of different sizes.
Using \method, we first make a prediction based only on the context set (without interactions). We then simulate corrections using either center clicks or centerline scribbles, and make additional predictions. For each example, we simulate interactive segmentation with 10 different random seeds and randomly sampled context sets of ground truth segmentation maps.

\subpara{Results}
Interactive segmentation performance improves as the size of the context set increases, demonstrating \method is able to use information from the context set to improve its predictions (\cref{fig:interseg_context_size}). There are diminishing returns to increasing the context set size. For example, performing another step of scribbles typically leads to a larger increase in Dice score, compared to doubling the context set size.

%--------------------------------------------------------------------------------------
\section{Conclusion}
%--------------------------------------------------------------------------------------

We presented \method, an interactive framework that enables rapid segmentation of an entire dataset of images, even for new tasks. \method leads to a substantial reduction of user interactions as more images are segmented. 

To enable \method, we introduce the first model that can perform \emph{interactive} segmentation of biomedical images \emph{in context}. The network segments an image given user interactions and a context set of previously labeled examples. Compared to ScribblePrompt, a state-of-the-art interactive segmentation model, \method reduces the number of clicks required to accurately segment a set of images, by 36\% on average on the first 18 images of a dataset.

\method opens up new opportunities for research into how best to prioritize sequentially segmenting images from a new dataset. Future research works could improve upon \method by investigating better context selection techniques~\cite{huo2024matchseg, gao2024boosting, wu2024efficient} to prioritize labeling images whose labels would be most informative for the segmentation task at hand.
\method has the potential to dramatically reduce the manual burden involved in segmenting datasets of biomedical images. 

%--------------------------------------------------------------------------------------

\section*{Acknowledgements}

This work was supported in part by Quanta Computer Inc. and the National Institute of Biomedical Imaging and Bioengineering of the National Institutes of Health under award number R01EB033773. Much of the computation required for this research was performed on computational hardware generously provided by the Massachusetts Life Sciences Center.

%--------------------------------------------------------------------------------------

{
    \small
    \bibliographystyle{ieeenat_fullname}
    \bibliography{main}
}
\clearpage
\maketitlesupplementary
\appendix

\MarkAppendixInTOC  
\setcounter{tocdepth}{2}
\AppendixOnlyTOC    

\section{Code}

Code and pre-trained weights are available at \href{https://multiverseg.csail.mit.edu}{https://multiverseg.csail.mit.edu}.

\section{\method Method}

\subsection{Architecture}

\subpara{CrossConv} We implement the CrossConvolutional layer slightly differently from ~\cite{universeg}. To avoid duplicate convolutions on the context features $v_i$ in Eq. 2, we partition weights $\theta_z$ channel-wise into $\{\theta_{z_1}, \theta_{z_2}\}$ and implement $z_i = LN(A(\text{Conv}(q, \theta_{z_1}) + \text{Conv}(v_i, \theta_{z_2})))$ where $q$ is the target feature map and $v_i$ is the feature map corresponding to context set entry $i$. We zero out the bias terms in $\text{Conv}(\cdot,\theta_{z_2})$ such that the computation is equivalent to $z_i = LN(A(\text{Conv}(q || v_i; \theta_z)))$. 

\subpara{Network}
We implement $f_\theta(\cdot)$ using an encoder with 5 encoder CrossBlock stages and a decoder with 4 CrossBlock stages. Each stage has 256 output features and LeakyReLU non-linearities after each convolution. We use bilinear interpolation for upsampling and downsampling.

The CrossBlock mechanism requires at least one context set entry. If the context set is empty, we use a dummy context set entry consisting of an image and segmentation with uniform value of 0.5. 

\section{Data}\label{appendix:data}

\subsection{Datasets}

We build on large dataset gathering efforts like MegaMedical~\cite{universeg, rakic2024tyche, wong2023scribbleprompt} to compile a collection of 79 open-access biomedical imaging datasets for training and evaluation, covering over 54k scans, 16 image types, and 713 labels.

\subpara{Division of Datasets}
The division of datasets and subjects for training, model selection, and evaluation is summarized in \cref{tab:data_split}. The 79 datasets were divided into 67 training datasets (\cref{tab:train_datasets} and 12 evaluation datasets (\cref{tab:eval_datasets}). Data from 9 (out of 12) of the evaluation datasets were used for model selection and final evaluation. The other 3 evaluation datasets were completely held-out from model selection and only used in the final evaluation. 

\subpara{Division of Subjects}
We split each dataset into 60\% train, 20\% validation, and 20\% test by subject. We used the ``train'' splits from the 67 training datasets to train \method models. We use the ``validation'' splits from the 67 training datasets and 9 validation datasets to select the best model checkpoint. We report final evaluation results across 12 held-out ``test'' splits of the 12 evaluation datasets to maximize the diversity of tasks and modalities in our evaluation set (\cref{tab:eval_datasets}). No data from the 9 validation datasets or 3 test datasets were seen by \method during training. 

\subpara{Task Definition}
We define a 2D segmentation task as a combination of (sub)dataset, axis (for 3D modalities), and label. For datasets with multiple segmentation labels, we consider each label separately as a binary segmentation task. For datasets with sub-datasets (e.g., malignant vs. benign lesions) we consider each cohort as a separate task. For multi-annotator datasets, we treat each annotator as a separate label. For instance segmentation datasets, we considered all instances as a single label. 

\subpara{3D Datasets}
For 3D modalities, we use the slice with maximum label area (``maxslice'') and the middle slice (``midslice'') for each volume for training of \method. For the 3D evaluation datasets (BTCV Cervix~\cite{BTCV}, ACDC~\cite{ACDC}, SCD~\cite{SCD}, SpineWeb~\cite{SpineWeb}, COBRE~\cite{COBRE}, TotalSegmentator~\cite{TotalSegmentator}) we evaluated the slice with the maximum label area for each subject, as in ~\cite{wong2023scribbleprompt}. We also considered evaluating on the middle slice, as in ~\cite{universeg, wu2024one,rakic2024tyche} and saw similar trends on the validation data. However, we opted for evaluation on maxslices because for our 3D test datasets (COBRE, TotalSegmentator) some labels do not appear in the midslices. Due to the large number of tasks in COBRE and TotalSegmentator, we only consider coronal slices from these datasets for evaluation. 

\subpara{Data Processing and Image Resolution}
We rescale image intensities to [0,1], padded square with zeros. For training, we resized images to $128^2$. In our final evaluations (\cref{sec:main_results}), we use images resized to $256^2$. We show additional evaluations on $128^2$ sized images in \cref{appendix:res128}.

\subpara{Data Sampling}
During training, we sample image, segmentation pairs hierarchically -- by dataset and modality, axis, and then label -- to balance training on datasets of different sizes.

\begin{table*}
    \footnotesize
    \centering
    \rowcolors{2}{white}{gray!15}
    \caption{\textbf{Dataset split overview}. Each dataset was split into 60\% train, 20\% validation and 20\% test by subject. Data from the ``train'' splits of the 67 training datasets were used to train the models. The \method models did not see any data from the validation datasets or test datasets during training. Data from the ``validation'' split of the 9 validation datasets was used for \method (\firstone{MVS}) model selection and experimenting with different evaluation methods of baselines. We report final results on the held-out test splits of 12 evaluation datasets: data from the ``test'' splits of the 9 validation datasets and the ``test'' splits of the 2 test datasets. To train the fully-supervised \secondone{nnUNet} baselines, we used the training and validation splits of the 12 evaluation datasets.
    }
    \label{tab:data_split}
    \resizebox{\textwidth}{!}{
    \begin{tabular}{cc||p{2.4cm}|p{4cm}|p{2cm}}
        & & \multicolumn{3}{c}{Split within each dataset by subject}
        \\
        \toprule
        Dataset Group & No. Datasets & {\centering Training Split (60\%)} & Validation Split (20\%) & Test Split (20\%)  \\
        \midrule
        Training Datasts & 67 & \firstone{MVS} training & \firstone{MVS} model selection & Not used  \\
        \hline
        Validation Datasets & 9 & \secondone{nnUNet} training & \firstone{MVS} and baselines model selection, \secondone{nnUNet} training & Final evaluation 
        \\ 
        \hline
        Test Datasets & 3 & \secondone{nnUNet} training & \secondone{nnUNet} training & Final evaluation 
        \\
        \bottomrule
    \end{tabular}
    }  
\end{table*}

\begin{table*}
\caption{
\textbf{Evaluation datasets}. We assembled the following set of datasets to evaluate \method and baseline methods. For the relative size of datasets, we include the number of unique scans (subject and modality pairs) and labels that each dataset has. These datasets were unseen by \method during training. \firstone{Three datasets} were completely held-out from model selection. The validation splits of the other 9 datasets were used for selecting the best model checkpoint. We report final results on the test splits of these 12 datasets.  
}
\label{tab:eval_datasets}
\centering
\rowcolors{2}{white}{gray!15}
\begin{tabular}{p{3.3cm}p{7cm}ccp{2.6cm}}
    \toprule 
    \textbf{Dataset Name} & \textbf{Description} & \textbf{Scans} & \textbf{Labels} & \textbf{Modalities} 
    \\ 
    \midrule
    \ ACDC~\cite{ACDC} & {Left and right ventricular endocardium} & 99 & 3 & cine-MRI 
    \\
    \ BTCV Cervix~\cite{BTCV} & Bladder, uterus, rectum, small bowel & 30 & 4 & CT  
    \\
    \ BUID~\cite{BUID}& Breast tumors & 647 & 2 & Ultrasound 
    \\
    \firstone{COBRE~\cite{COBRE,fischl2012freesurfer,neurite}} & Brain anatomy & 258 & 45 & T1-weighted MRI
    \\
    \ DRIVE~\cite{DRIVE} & Blood vessels in retinal images & 20 & 1 & Optical camera
    \\
    \ HipXRay~\cite{HipXRay} & Ilium and femur & 140 & 2 & X-Ray 
    \\
    \ PanDental~\cite{PanDental} & Mandible and teeth & 215 & 2 & X-Ray 
    \\
    \ SCD~\cite{SCD} & Sunnybrook Cardiac Multi-Dataset Collection & 100 & 1 & cine-MRI 
    \\
    \firstone{SCR~\cite{SCR}} & Lungs, heart, and clavicles  & 247 & 5 & X-Ray
    \\
    \ SpineWeb~\cite{SpineWeb} & Vertebrae & 15 & 1 & T2-weighted MRI  
    \\
    \firstone{TotalSegmentator~\cite{TotalSegmentator}} & 104 anatomic structures (27 organs, 59 bones, 10 muscles, and 8 vessels) & 1,204 & 104 & CT 
    \\
    \ WBC~\cite{WBC} & White blood cell cytoplasm and nucleus & 400 & 2 & Microscopy 
    \\
    \bottomrule
\end{tabular}
\end{table*}

\begin{table*}

\caption{\textbf{Train datasets}. We train \method on the following datasets. For the relative size of datasets, we have included the number of unique scans (subject and modality pairs) that each dataset has.}
\label{tab:train_datasets}
\centering
\rowcolors{2}{white}{gray!15}
\resizebox{\textwidth}{!}{
\begin{tabular}{p{3.8cm}p{9.5cm}cp{4cm}}
    \textbf{Dataset Name} & \textbf{Description} & \textbf{Scans} & \textbf{Modalities} 
    \\ 
    \toprule 
     AbdominalUS~\cite{AbdominalUS} & Abdominal organ segmentation & 1,543 & Ultrasound
     \\
     AMOS~\cite{AMOS} & Abdominal organ segmentation & 240 & CT, MRI 
     \\
     BBBC003~\cite{BBBC003} & Mouse embryos & 15 & Microscopy 
     \\
     BBBC038~\cite{BBBC038} & Nuclei instance segmentation & 670 & Microscopy 
     \\
     BrainDev~\cite{gousias2012magnetic, BrainDevFetal, BrainDevelopment, serag2012construction} & Adult and neonatal brain atlases & 53 & Multimodal MRI 
     \\
     BrainMetShare\cite{BrainMetShare} & Brain tumors & 420 & Multimodal MRI 
     \\
     BRATS~\cite{BRATS, bakas2017advancing, menze2014multimodal} & Brain tumors & 6,096 & Multimodal MRI 
     \\
     BTCV Abdominal~\cite{BTCV} & 13 abdominal organs & 30 & CT 
     \\  
     BUSIS~\cite{BUSIS} & Breast tumors & 163 & Ultrasound 
     \\
     CAMUS~\cite{CAMUS}  & Four-chamber and Apical two-chamber heart & 500 & Ultrasound
     \\
     CDemris~\cite{cDemris}  & Human left atrial wall & 60 & CMR 
     \\
     CHAOS~\cite{CHAOS_1, CHAOS_2}  & Abdominal organs (liver, kidneys, spleen) & 40 & CT, T2-weighted MRI 
     \\
     CheXplanation~\cite{CheXplanation} & Chest X-Ray observations & 170 & X-Ray
     \\
     CoNSeP & Histopathology Nuclei & 27 & Microscopy
     \\
     CT2US~\cite{CT2US} & Liver segmentation in synthetic ultrasound & 4,586 & Ultrasound
     \\
     CT-ORG\cite{CT_ORG} & Abdominal organ segmentation (overlap with LiTS) & 140 & CT 
     \\
     DDTI~\cite{DDTI} & Thyroid segmentation & 472 & Ultrasound
     \\
     DukeLiver~\cite{DukeLiver} & Liver segmentation in abdominal MRI & 310 & MRI
     \\
     EOphtha~\cite{EOphtha} & Eye microaneurysms and diabetic retinopathy & 102 & Optical camera
     \\
     FeTA~\cite{FeTA} & Fetal brain structures & 80 & Fetal MRI 
     \\
     FetoPlac~\cite{FetoPlac} & Placenta vessel & 6 & Fetoscopic optical camera
     \\
     FLARE~\cite{FLARE21} & Abdominal organs (liver, kidney, spleen, pancreas) & 361 & CT 
     \\
     HaN-Seg~\cite{HaN-Seg} & Head and neck organs at risk & 84 & CT, T1-weighted MRI
     \\
     HMC-QU~\cite{HMC-QU, kiranyaz2020left} & 4-chamber (A4C) and apical 2-chamber (A2C) left  wall & 292 & Ultrasound 
     \\ 
     I2CVB~\cite{I2CVB} & Prostate (peripheral zone, central gland) & 19 & T2-weighted MRI 
     \\
     IDRID~\cite{IDRID} & Diabetic retinopathy & 54 & Optical camera 
     \\
     ISBI-EM~\cite{ISBI_EM} & Neuronal structures in electron microscopy & 30 & Microscopy
     \\
     ISIC~\cite{ISIC} & Demoscopic lesions & 2,000 & Dermatology
     \\
     ISLES~\cite{ISLES} & Ischemic stroke lesion & 180 & Multimodal MRI 
     \\
     KiTS~\cite{KiTS} & Kidney and kidney tumor & 210 & CT 
     \\
     LGGFlair~\cite{buda2019association, LGGFlair} & TCIA lower-grade glioma brain tumor & 110 & MRI 
     \\
     LiTS~\cite{LiTS} & Liver tumor & 131 & CT 
     \\
     LUNA~\cite{LUNA} & Lungs & 888 & CT 
     \\
     MCIC~\cite{MCIC} & Multi-site brain regions of schizophrenic patients & 390 & T1-weighted MRI
     \\
     MMOTU~\cite{MMOTU} & Ovarian tumors & 1,140 & Ultrasound
     \\
     MSD~\cite{MSD} & Large-scale collection of 10 medical segmentation datasets & 3,225 & CT, Multimodal MRI
     \\
     MuscleUS~\cite{MuscleUS} & Muscle segmentation (biceps and lower leg) & 8,169 & Ultrasound
     \\
     NCI-ISBI~\cite{NCI-ISBI} & Prostate & 30 & T2-weighted MRI 
     \\
     NerveUS~\cite{NerveUS} & Nerve segmentation & 5,635 & Ultrasound
     \\
     OASIS~\cite{OASIS-proc, OASIS-data} & Brain anatomy & 414 & T1-weighted MRI 
     \\
     OCTA500~\cite{OCTA500} & Retinal vascular & 500 & OCT/OCTA 
     \\
     PanNuke~\cite{PanNuke} & Nuclei instance segmentation & 7,901 & Microscopy
     \\
     PAXRay~\cite{PAXRay} & 92 labels covering lungs, mediastinum, bones, and sub-diaphram in Chest X-Ray & 852 & X-Ray 
     \\
    PROMISE12~\cite{Promise12} & Prostate & 37 & T2-weighted MRI
    \\
    PPMI~\cite{PPMI,dalca2018anatomical} & Brain regions of Parkinson patients & 1,130 & T1-weighted MRI
    \\
    QUBIQ~\cite{qubiq} & Collection of 4 multi-annotator datasets (brain, kidney, pancreas and prostate) & 209 & T1-weighted MRI, Multimodal MRI, CT
    \\
     ROSE~\cite{Rose} & Retinal vessel & 117 & OCT/OCTA 
     \\
     SegTHOR~\cite{SegTHOR} & Thoracic organs (heart, trachea, esophagus) & 40 & CT 
     \\
     SegThy~\cite{SegThy} & Thyroid and neck segmentation & 532 & MRI, Ultrasound
     \\
     ssTEM~\cite{ssTEM} & Neuron membranes, mitochondria, synapses and extracellular space & 20 & Microscopy
     \\
     STARE~\cite{STARE} & Blood vessels in retinal images & 20 & Optical camera 
     \\
     ToothSeg~\cite{ToothSeg} & Individual teeth & 598 & X-Ray
     \\
     VerSe~\cite{VerSe} & Individual vertebrae & 55 & CT
     \\
     WMH~\cite{WMH} & White matter hyper-intensities & 60 & Multimodal MRI 
     \\
     WORD~\cite{Word} & Abdominal organ segmentation & 120 & CT 
     \\
     \bottomrule
\end{tabular}
}
\end{table*}

\subsection{Synthetic Task Generation}
\label{appendix:synth_aug}

We introduce a new approach for constructing synthetic tasks from real images. Given a single image $x_0$, we construct a set of images $\{x_i', y_i'\}_{i=1}^{m+1}$ representing a synthetic task. We then partition this set into a target example and context set of size $m$ for training. 

\subpara{Related Work}
Although previous work found that training on a mix of real and synthetic segmentation \emph{labels} based on image superpixels is useful for improving generalization in interactive segmentation~\cite{wong2023scribbleprompt}, we do not use such data here. That approach cannot be directly applied to \method because it does not produce semantically consistent labels across multiple images.

\begin{figure*}
    \centering
    \includegraphics[width=\linewidth]{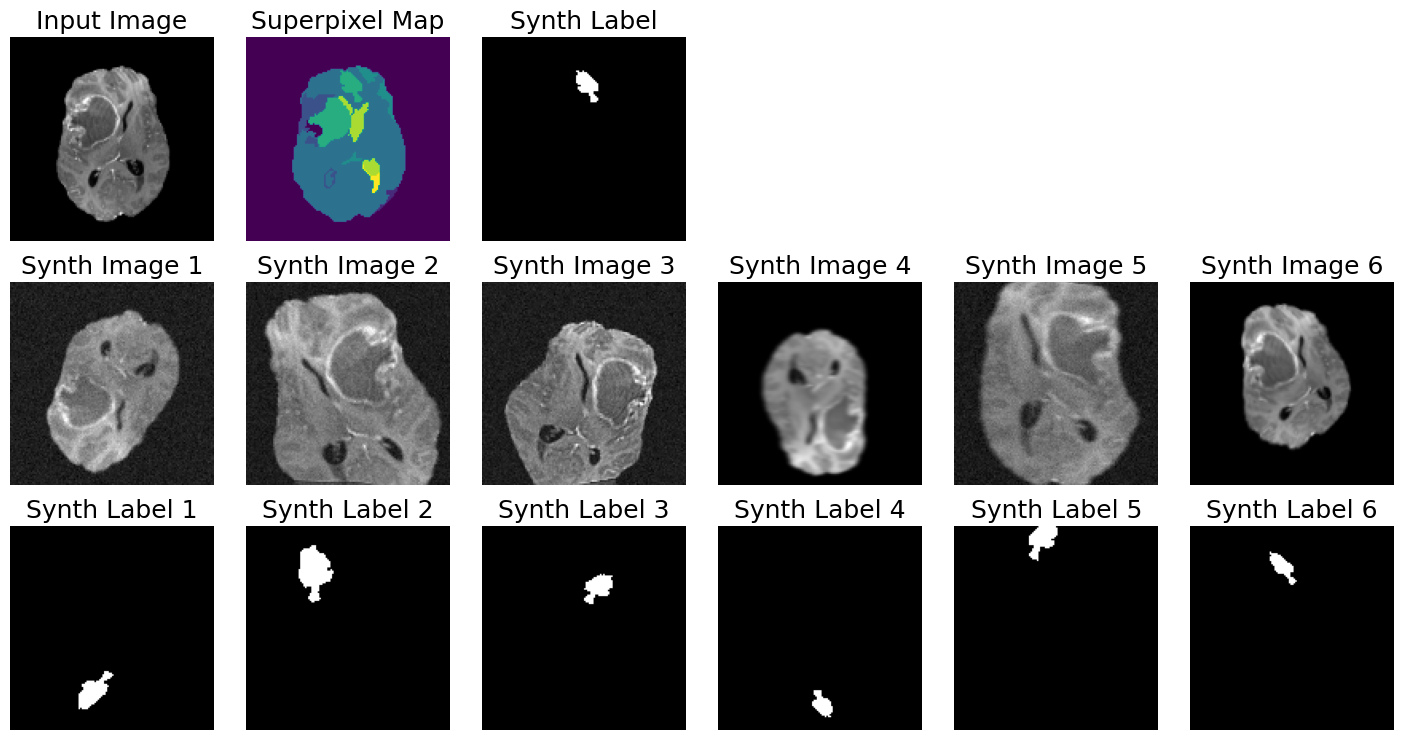}
    \caption{\textbf{Synthetic task generation example}. Given an input image, we apply a superpixel algorithm to generate a superpixel map of potential synthetic labels. We randomly sample one of the superpixels to serve as a synthetic label. Next, we duplicate the input image and synthetic label $m+1$ times and apply data augmentations (\cref{tab:synth_augmentations}) to vary the examples within the synthetic task. We use the first synthetic example as the target and the remaining $m$ synthetic examples as the context set during training.}
    \label{fig:synth_example}
\end{figure*}

\subpara{Method}
To build a synthetic task from an image, we first generate a synthetic label and then perform aggressive augmentations to create a set of images corresponding to the same synthetic task (\cref{fig:synth_example}).

Given an image $x_0$, we first generate a synthetic label $y_{synth}$ by applying a superpixel algorithm \cite{felzenszwalb_superpixels} with scale parameter $\lambda \sim U[1,\lambda_{max}]$ to partition the image into a multi-label mask of $k$ superpixels $z \in \{1, \dots, k\}^{n \times n}$. We then randomly select a superpixel $y_{synth} = \mathbbm{1} (z = c)$ as a synthetic label. 

To generate a set of $m+1$ images representing the same task, we duplicate $(x_0, y_{synth})$, $m+1$ times and apply aggressive augmentations to vary the images and segmentation labels~\cite{zhao2019dataaug,universeg}.

\subpara{Implementation} 
\method was trained with $p_{synth}=0.5$. We use a superpixel algorithm~\cite{felzenszwalb_superpixels} with $\lambda \sim [1,500]$. \cref{tab:synth_augmentations} lists the data augmentations.

\begin{table}
    \centering
    \rowcolors{2}{white}{gray!15}
    \begin{tabular}{lcc}
    \toprule
    \textbf{Augmentations} & $p$ & Parameters
    \\
    \midrule
    % Random Affine
    \cellcolor{gray!15}  &\cellcolor{gray!15} & \cellcolor{gray!15} $\text{degrees} \in [-25, 25]$ 
    \\
    \cellcolor{gray!15}    & \cellcolor{gray!15} & \cellcolor{gray!15} $\text{translation} \in [0, 0.2]$ 
    \\
    \multirow{-3}{*}{\cellcolor{gray!15}Random Affine} & \multirow{-3}{*}{\cellcolor{gray!15} 0.8} & \cellcolor{gray!15} $\text{scale} \in [0.9, 1.5]$
    \\
    %Brightness Contrast 
    \cellcolor{white}  & \cellcolor{white} & \cellcolor{white} $\text{brightness}\in[-0.1, 0.1]$
    \\
    \multirow{-2}{*}{\cellcolor{white}{Brightness Contrast}} &  \multirow{-2}{*}{\cellcolor{white}{0.5}} & \cellcolor{white} $\text{contrast}\in[0.5, 1.5]$ 
    \\
    % Elastic Transform
    \cellcolor{gray!15} & \cellcolor{gray!15} & \cellcolor{gray!15} $\alpha\in[1, 10]$
    \\
    \multirow{-2}{*}{\cellcolor{gray!15}Elastic Transform} & \multirow{-2}{*}{\cellcolor{gray!15} 0.8} & \cellcolor{gray!15} $\sigma\in[8,15]$
    \\
    % Sharpness
    Sharpness & 0.5 & $\text{sharpness} = 5$
    \\
    %Gaussian Blur
    \cellcolor{gray!15} & \cellcolor{gray!15} & \cellcolor{gray!15} $\sigma\in[0.1, 1.5]$ 
    \\
    \multirow{-2}{*}{\cellcolor{gray!15}Gaussian Blur} & \multirow{-2}{*}{\cellcolor{gray!15} 0.5} & \cellcolor{gray!15} $k=5$
    \\
    % Gaussian Noise
    \cellcolor{white} & \cellcolor{white} & \cellcolor{white} $\mu\in[0, 0.05]$
    \\
    \multirow{-2}{*}{\cellcolor{white}Gaussian Noise} & \multirow{-2}{*}{\cellcolor{white} 0.5} & \cellcolor{white} $ \sigma \in [0, 0.05]$
    \\
    % Horizontal Flip
    Horizontal Flip & 0.5 & None 
    \\
    % Vertical Flip
    Vertical Flip & 0.5 & None 
    \\
    \bottomrule
    \end{tabular}
    \caption{\textbf{Data augmentations for generating synthetic tasks}. Given a set of $m+1$ copies of the same example, we randomly sampled data augmentations for each instance to increase the diversity of examples within the task. Each augmentation is sampled with probability $p$.}
    \label{tab:synth_augmentations}
\end{table}

\subsection{Data Augmentation}
\label{appendix:data_aug}

\cref{tab:augmentations} shows the within-task augmentations and task-augmentations used to train \method~\cite{universeg,rakic2024tyche}.

\begin{table}
\begin{subtable}[h]{\linewidth}
    \centering
    \rowcolors{2}{white}{gray!15}
    \begin{tabular}{lcc}
    \toprule
   \textbf{Augmentations} & $p$ & Parameters\\
    \midrule
    % Random Affine
    \cellcolor{gray!15}  &\cellcolor{gray!15} & \cellcolor{gray!15} $\text{degrees} \in [-25, 25]$ 
    \\
    \cellcolor{gray!15}    & \cellcolor{gray!15} & \cellcolor{gray!15} $\text{translation} \in [0, 0.1]$ 
    \\
    \multirow{-3}{*}{\cellcolor{gray!15}Random Affine} & \multirow{-3}{*}{\cellcolor{gray!15} 0.25} & \cellcolor{gray!15} $\text{scale} \in [0.9, 1.1]$
    \\
    %Brightness Contrast 
    \cellcolor{white}  & \cellcolor{white} & \cellcolor{white} $\text{brightness}\in[-0.1, 0.1]$
    \\
    \multirow{-2}{*}{\cellcolor{white}{Brightness Contrast}} & \multirow{-2}{*}{\cellcolor{white}{0.25}} & \cellcolor{white} $\text{contrast}\in[0.5, 1.5]$ 
    \\
    % Elastic Transform
    \cellcolor{gray!15} & \cellcolor{gray!15} & \cellcolor{gray!15}$\alpha\in[1, 2.5]$
    \\
    \multirow{-2}{*}{\cellcolor{gray!15}Elastic Transform} & \multirow{-2}{*}{\cellcolor{gray!15} 0.8} & \cellcolor{gray!15} $\sigma\in[7,9]$
    \\
    % Sharpness
    Sharpness & 0.25 & $\text{sharpness} = 5$
    \\
    %Gaussian Blur
    \cellcolor{gray!15} & \cellcolor{gray!15} & \cellcolor{gray!15} $\sigma\in[0.1, 1.0]$ 
    \\
    \multirow{-2}{*}{\cellcolor{gray!15}Gaussian Blur} & \multirow{-2}{*}{\cellcolor{gray!15} 0.25} & \cellcolor{gray!15} $k=5$
    \\
    % Gaussian Noise
    \cellcolor{white} & \cellcolor{white} & \cellcolor{white} $\mu\in[0, 0.05]$
    \\
    \multirow{-2}{*}{\cellcolor{white}Gaussian Noise} & \multirow{-2}{*}{\cellcolor{white} 0.25} &  \cellcolor{white} $ \sigma \in [0, 0.05]$
    \\
    \bottomrule
    \end{tabular}
    \caption{\textbf{Within-Task Augmentations\vspace{1em}}}
\end{subtable}
~
\begin{subtable}[h]{\linewidth}
\centering
\rowcolors{2}{white}{gray!15}
    \begin{tabular}{lcc}
    \toprule
    \textbf{Augmentations} & $p$ & Parameters\\
    \midrule
    % Random Affine
    &  &  $\text{degrees}\in[0, 360]$ 
    \\
    \cellcolor{gray!15} & \cellcolor{gray!15} & \cellcolor{gray!15} $\text{translates}\in[0, 0.2]$ 
    \\
    \multirow{-3}{*}{Random Affine} & \multirow{-3}{*}{ 0.5} & $\text{scale} \in [0.8, 1.1]$
    \\
    %Brightness Contrast 
    &  & $\text{brightness} \in [-0.1, 0.1]$ 
    \\
    \multirow{-2}{*}{\cellcolor{white}{Brightness Contrast}} &  \multirow{-2}{*}{\cellcolor{white}{0.5}} & \cellcolor{white} $\text{contrast} \in [0.8, 1.2]$ 
    \\
    %Gaussian Blur
    \cellcolor{gray!15} &  \cellcolor{gray!15} & \cellcolor{gray!15}$\sigma\in[0.1, 1.1]$ 
    \\
    \multirow{-2}{*}{\cellcolor{gray!15}Gaussian Blur}  & \multirow{-2}{*}{\cellcolor{gray!15} 0.5} & \cellcolor{gray!15} $k=5$ 
    \\
    % Gaussian Noise
    \cellcolor{white} & \cellcolor{white} & \cellcolor{white} $\mu\in[0, 0.05]$
    \\
    \multirow{-2}{*}{\cellcolor{white}Gaussian Noise} & \multirow{-2}{*}{\cellcolor{white} 0.5} & \cellcolor{white} $ \sigma \in [0, 0.05]$\\
    % Elastic Transform
    \cellcolor{gray!15}  & \cellcolor{gray!15} & \cellcolor{gray!15} $\alpha\in[1, 2]$
    \\
    \multirow{-2}{*}{\cellcolor{gray!15}Elastic Transform} & \multirow{-2}{*}{\cellcolor{gray!15} 0.5} & \cellcolor{gray!15} $\sigma\in[6,8]$
    \\
    % Sharpness
    \cellcolor{white}Sharpness & \cellcolor{white}0.5 & \cellcolor{white} $\text{sharpness}=5$
    \\
    % Horizontal Flip
    Horizontal Flip &  0.5 & None 
    \\
    % Vertical Flip
    Vertical Flip & 0.5 & None 
    \\
    % Vertical Flip
    Sobel Edges Label & 0.5 & None
    \\
    % Intensities
    Flip Intensities & 0.5 & None
    \\\hline
    \end{tabular}
    \caption{\textbf{Task Augmentations}}
    \end{subtable}
    \caption{\textbf{Augmentations used to train \method.} Within-task data augmentations (\textbf{top}) are randomly sampled for each example within a task to increase the diversity within a task. Task augmentations (\textbf{bottom}) are randomly sampled for each task and then applied to all examples in a task to increase the diversity of tasks. Each augmentation is randomly sampled with probability $p$. We apply augmentations after (optional) synthetic task generation and before simulating user interactions.}
    \label{tab:augmentations}
\end{table}

%----------------------------------------------------------------
\section{Experimental Setup}

\subsection{Baselines}
\label{appendix:baselines}

We provide additional details on the baselines. We summarize the capabilities of our method and baselines in \cref{tab:baselines}.

\begin{table*}
    \centering
    \rowcolors{2}{white}{gray!15}
    \begin{tabular}{l|c|c|c}
        \toprule
         Method & Interactive & In-Context & Interactive In-Context \\
         \midrule 
         SAM~\cite{SAM} & $\checkmark$ & &
         \\
         MedSAM~\cite{MedSAM} & $\checkmark$ & &
         \\
         SAM-Med2D~\cite{cheng_sam-med2d_2023} & $\checkmark$ & &
         \\
         SegNext~\cite{liu2024rethinking} & $\checkmark$ & &
         \\  ScribblePrompt~\cite{wong2023scribbleprompt} &
         $\checkmark$ & & \\
         \midrule
         UniverSeg~\cite{universeg} & & $\checkmark$ & \\
         LabelAnything~\cite{LabelAnything} & & $\checkmark$ & \\
         \midrule
         OnePrompt~\cite{wu2024one} & $\checkmark$ & $\checkmark$ (context size = 1) & \\
         SP+UVS & $\checkmark$ & $\checkmark$ & $\checkmark$ \\
         \midrule
         MultiverSeg (ours) & $\checkmark$ & $\checkmark$ & $\checkmark$
         \\
         \bottomrule
    \end{tabular}
    \caption{\textbf{Summary of segmentation methods}.}
    \label{tab:baselines}
\end{table*}

\subpara{SAM}
We evaluated SAM~\cite{SAM} (ViT-b) in both ``single-mask'' and ``multi-mask'' mode on our validation data, and average results were better using ``single-mask'' mode. We report final results for SAM on the test data using ``single-mask'' mode. 

\subpara{UniverSeg} Previous work found that ensembling UniverSeg predictions across multiple randomly sampled context sets improved Dice score~\cite{universeg}. We report results \emph{without} ensembling to accurately reflect the mean Dice of predictions given a fixed size context set.

\subpara{OnePrompt} OnePrompt~\cite{wu2024one} is a medical image segmentation model that can perform in-context segmentation of a target image given a single context example with scribble, click, bounding box or mask annotation on the context image. OnePrompt can also be used for interactive segmentation by using the same image as both the context image and the target image. We do not compare to OnePrompt because the pre-trained model weights are not publicly available. Recreating the data processing and retraining the model was beyond our computational capacity. For reference, the OnePrompt model required 64 NVIDIA A100 GPUs to train~\cite{wu2024one}.

\subpara{LabelAnything} LabelAnything~\cite{LabelAnything} is an in-context segmentation model designed for few-shot multi-label segmentation of natural images. LabelAnything takes as input a target image to segment and a context set of images with multi-label mask, click, or bounding box annotations. We do not compare to LabelAnything because the pre-trained model weights are not publicly available. As with OnePrompt, recreating the data handling and retraining the model from scratch was beyond our computational capacity.

\subsection{Inference}

\subpara{Image Resolution}
\method, ScribblePrompt, and UniverSeg, which were all developed and trained on $128^2$ sized images, and output predictions at the same resolution. SAM was trained with $1024^2$ sized inputs and predicts segmentations at $256^2$ resolution. 
For each method, we resized the inputs to the method's training input size using bilinear interpolation before performing inference and then resized the output (as needed) to the evaluation resolution. 

\subsection{Metrics}

\subpara{Averaging} When reporting average performance for a dataset or across multiple datasets, we averaged metrics hierarchically by subject, label, axis, modality, subdataset, and then dataset. 

\subpara{Confidence Intervals} 
For Experiment 1, we calculate 95\% confidence intervals over results from 200 simulations with different random seeds. For Experiment 2, we calculate 95\% confidence intervals by bootstrapping over subjects with 100 runs.

%---------------------------------------------------------
\section{Experiment 1: Evaluation}
%---------------------------------------------------------

\subsection{Setup}
\label{appendix:experiment1_setup}

\begin{figure*}
    \centering
    \includegraphics[width=\linewidth]{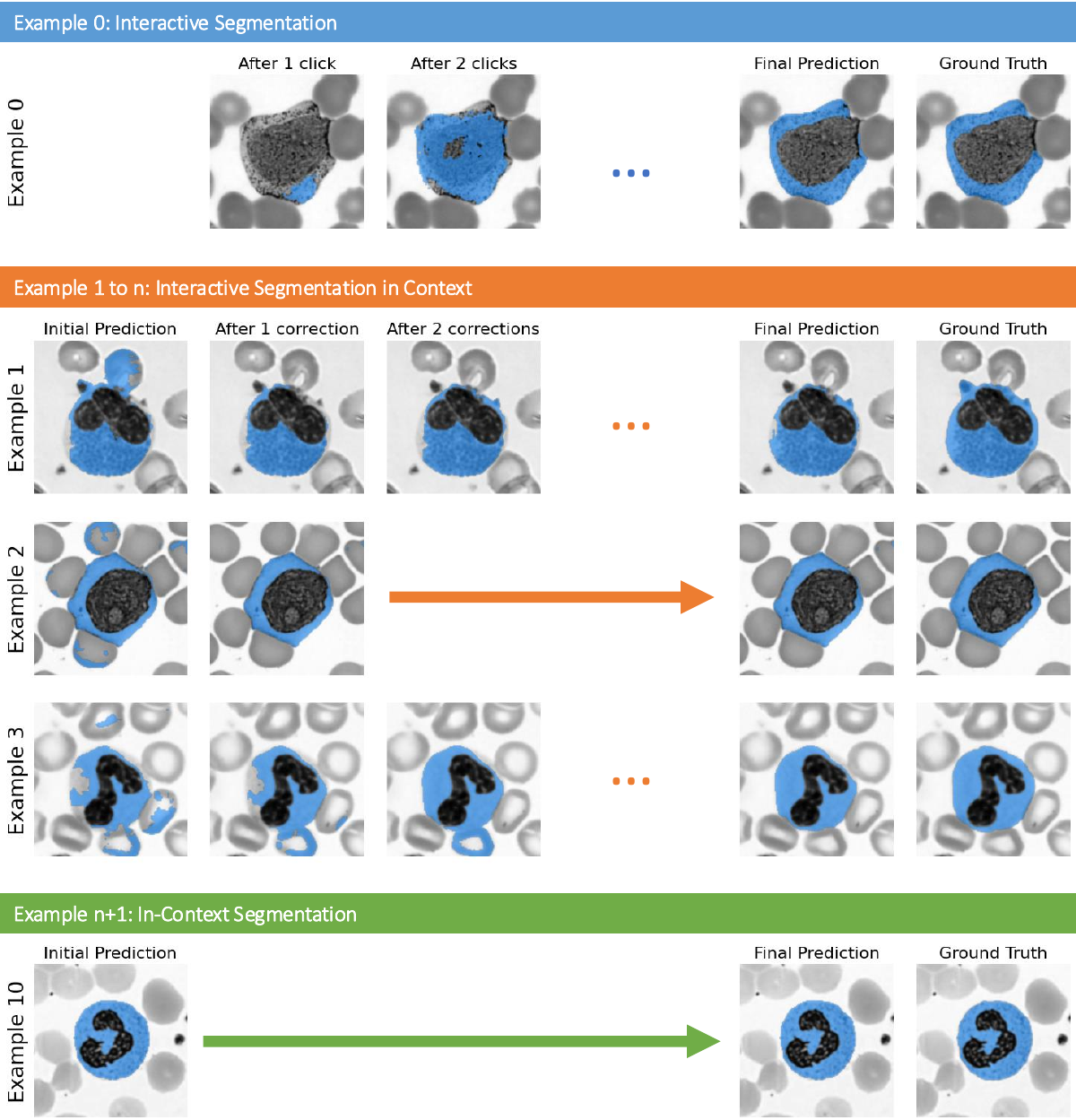}
    \caption{
    \textbf{Example segmentation process with \method}. 
    We begin by interactively segmenting a seed image (\textbf{Example 0}) to 90\% Dice. The Example 0 image and final prediction are added to the context set for subsequent examples.
    For each subsequent example, we first make an initial in-context segmentation prediction using a context set containing all the previous examples and previously predicted segmentations. Then, we simulate center correction clicks until the predicted segmentation achieves $\geq 90\%$ Dice or we have accrued 20 clicks.
    For \textbf{Example 2}, we only simulated 1 correction because the prediction reached $90\%$ Dice after 1 correction click. For \textbf{Example 1} and \textbf{Example 3}, additional correction clicks were needed.
    When the context set is large enough ($>$$n$), the in-context prediction from \method may be accurate enough that no corrections are needed. For \textbf{Example 10}, the Dice score of the predicted in-context segmentation is greater than $90\%$ so we do not need to simulate any corrections. In practice, $n$ varies by task.
    }
    \label{fig:evaluation_examples}
\end{figure*}

We illustrate the process of segmenting a set of images using \method in \cref{fig:evaluation_examples}

\subpara{Procedure}
For all methods, we interactively segment a seed image to 90\% Dice using ScribblePrompt. This first image was randomly sampled (for each simulation round) from the training split. Since the number of interactions and the prediction for this seed image is the same for all methods, we exclude it from the reported results.

We report the number of interactions to achieve 90\% Dice for each of the \emph{next} 18 images from the held-out test split of our evaluation tasks. We conduct 200 rounds of simulations, randomly sampling 18 test images (without replacement) from each task and sequentially segmenting them using each method. We use the same random seeds for each method, so the sampled examples are the same across methods for each simulation round.

\begin{figure}
    \centering
    \includegraphics[width=\linewidth]{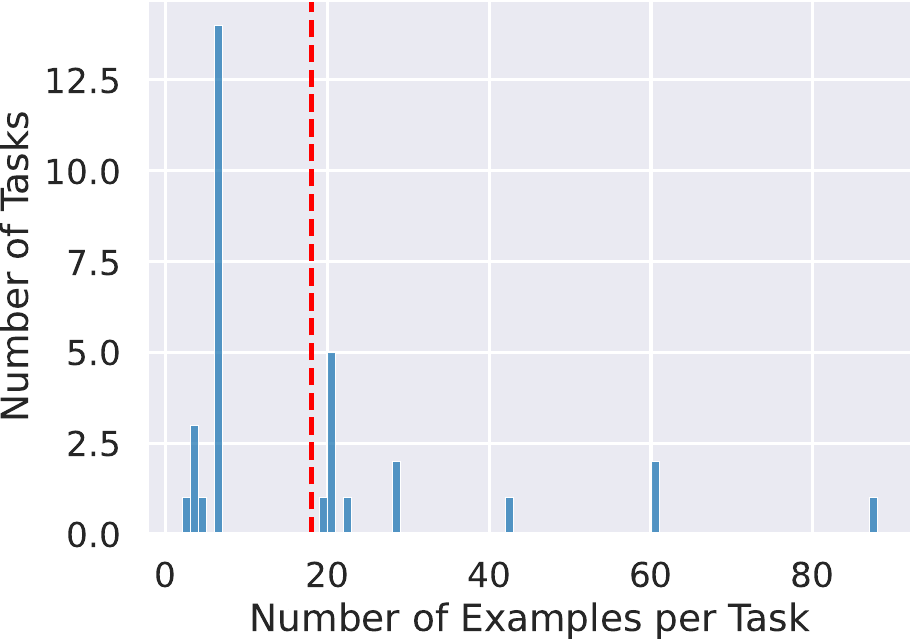}
    \caption{\textbf{Examples per task}. We visualize the distribution of examples per task in our validation data. We only consider tasks with at least 18 examples in Experiment 1.}
    \label{fig:task_size_distribution}
\end{figure}

\subpara{Tasks}
We exclude tasks with fewer than 18 test examples, leaving 161 tasks from 8 evaluation datasets~\cite{COBRE,ACDC,BUID,PanDental,HipXRay,SCR,WBC,TotalSegmentator}. We selected this cutoff based on the distribution of task sizes in our validation data (\cref{fig:task_size_distribution}) to focus on scenarios where a user wants to segment many similar images.

\subpara{Data}
We conducted our evaluation on $256^2$ sized images. For each method, we resized the inputs to match the size of the model's training data before performing the forward pass, and then resized the prediction back to $256^2$ before calculating the Dice Score. In \cref{appendix:res128} we conduct a sensitivity analysis, performing the evaluation with $128^2$ sized images

\subsection{Interactions per Image as a Function of Dataset Size}
\label{appendix:interactions_per_example}

\begin{figure*}
    \centering
    \includegraphics[width=0.75\linewidth]{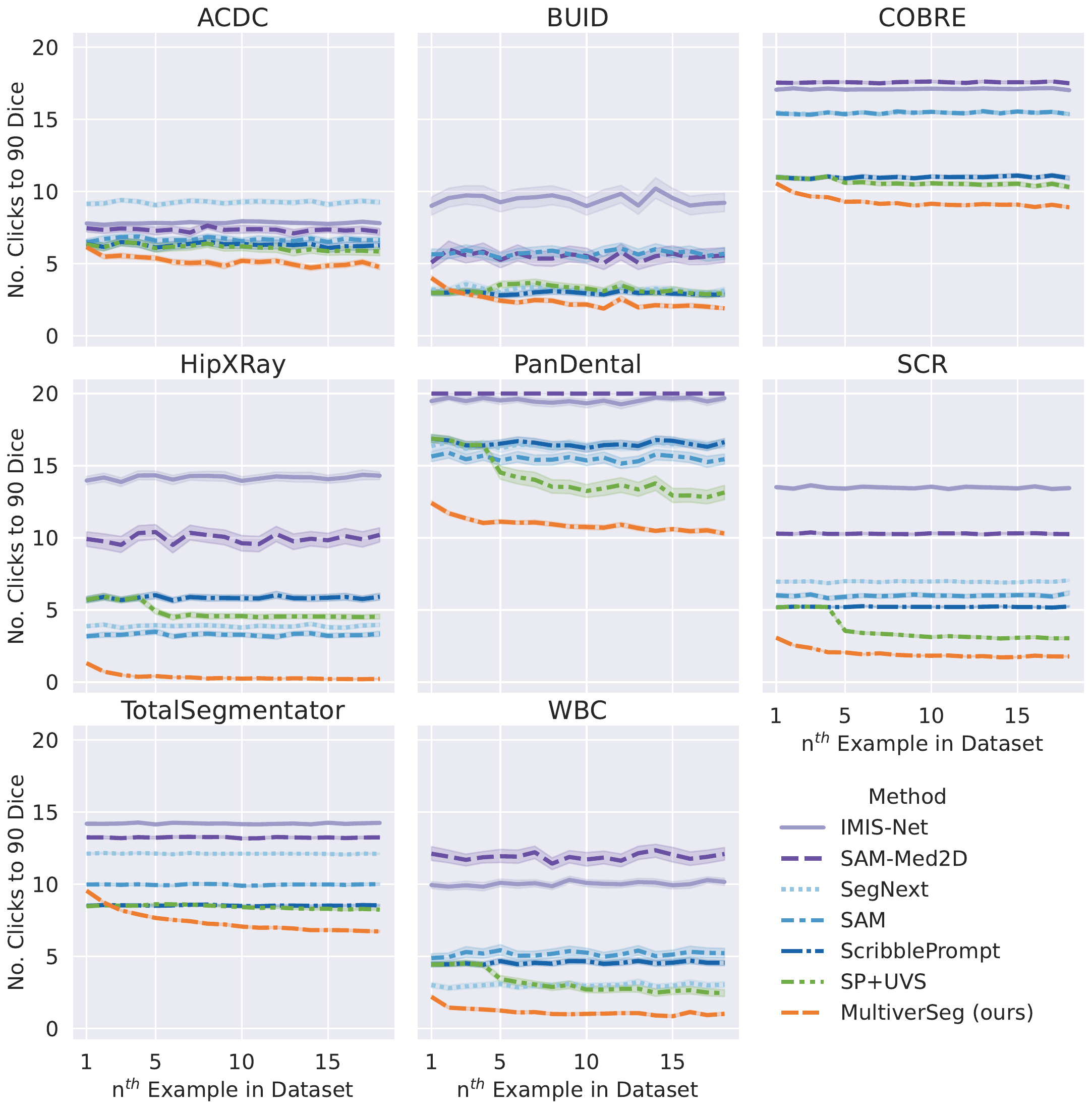}
    \caption{\textbf{Clicks to target Dice on unseen datasets}. Number of interactions needed to reach 90\% Dice as a function of the example number being segmented. For the $n^{th}$ image being segmented, the context set has $n$ examples. \method requires substantially fewer interactions to achieve 90 Dice than the baselines, and as more images are segmented, the average number of interactions required decreases dramatically. Shaded regions show 95\% CI from bootstrapping.}
    \label{fig:clicks_to_dice_dataset}
\end{figure*}

\begin{figure*}
    \centering
    \includegraphics[width=0.75\linewidth]{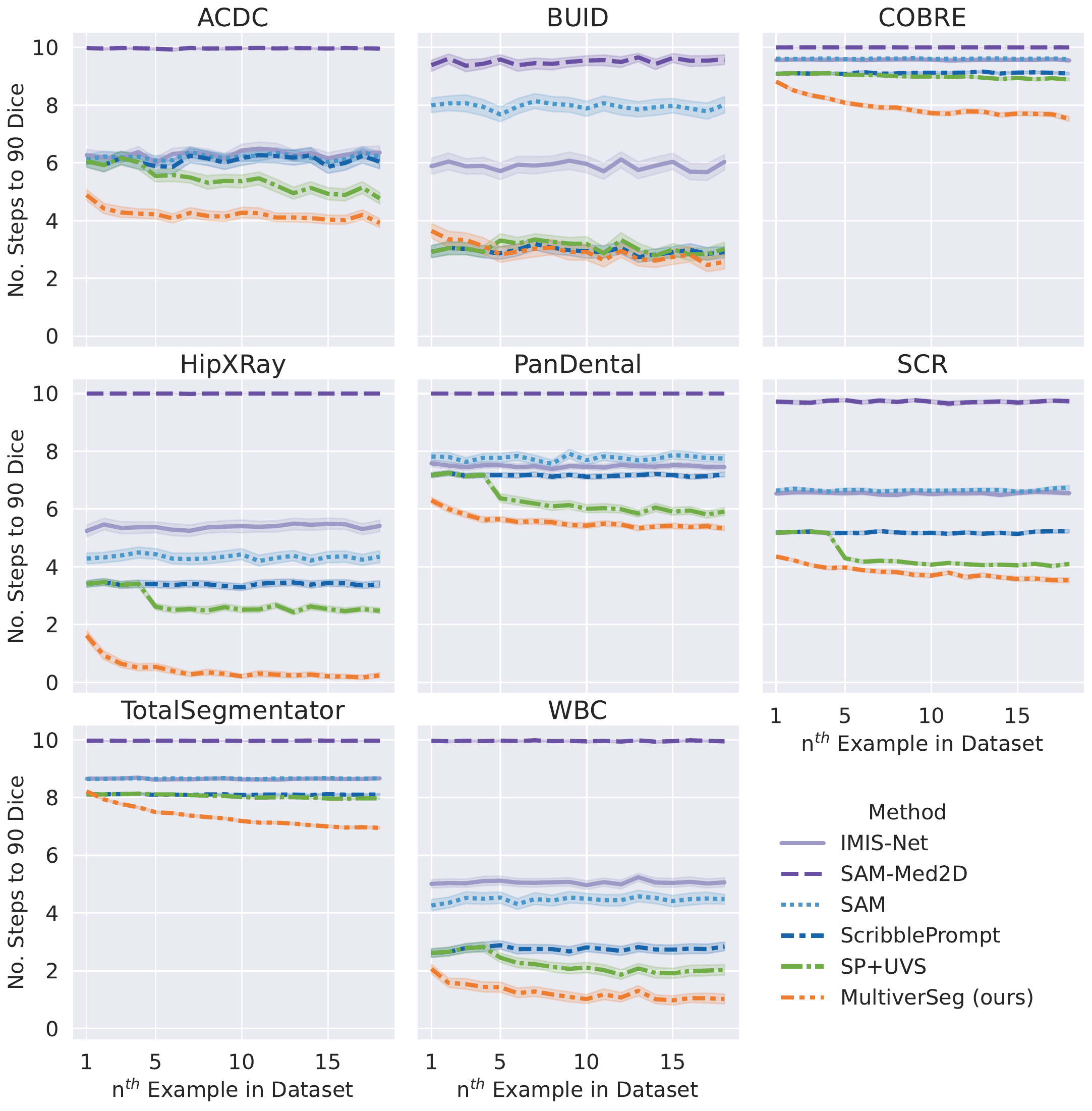}
    \caption{\textbf{Scribbles to target Dice on unseen datasets}. Number of interactions needed to reach 90\% Dice as a function of the example number being segmented. For the $n^{th}$ image being segmented, the context set has $n$ examples. \method requires substantially fewer interactions to achieve 90 Dice than the baselines, and as more images are segmented, the average number of interactions required decreases dramatically. Shaded regions show 95\% CI from bootstrapping.}
    \label{fig:scribbles_to_dice_dataset}
\end{figure*}

\subpara{Results by dataset}
As more examples are segmented and the context set grows, the number of clicks and scribbles required to get to $90\%$ Dice on the $\text{n}^\text{th}$ example using \method decreases substantially. \cref{fig:clicks_to_dice_dataset} and \cref{fig:scribbles_to_dice_dataset} show results averaged by dataset. \method and SP+UVS are less effective at reducing the number of clicks for tasks from BUID, a breast ultrasound lesion segmentation dataset, perhaps due to the heterogeneity of examples in that dataset. 

\begin{figure}
    \centering
    \includegraphics[width=\linewidth]{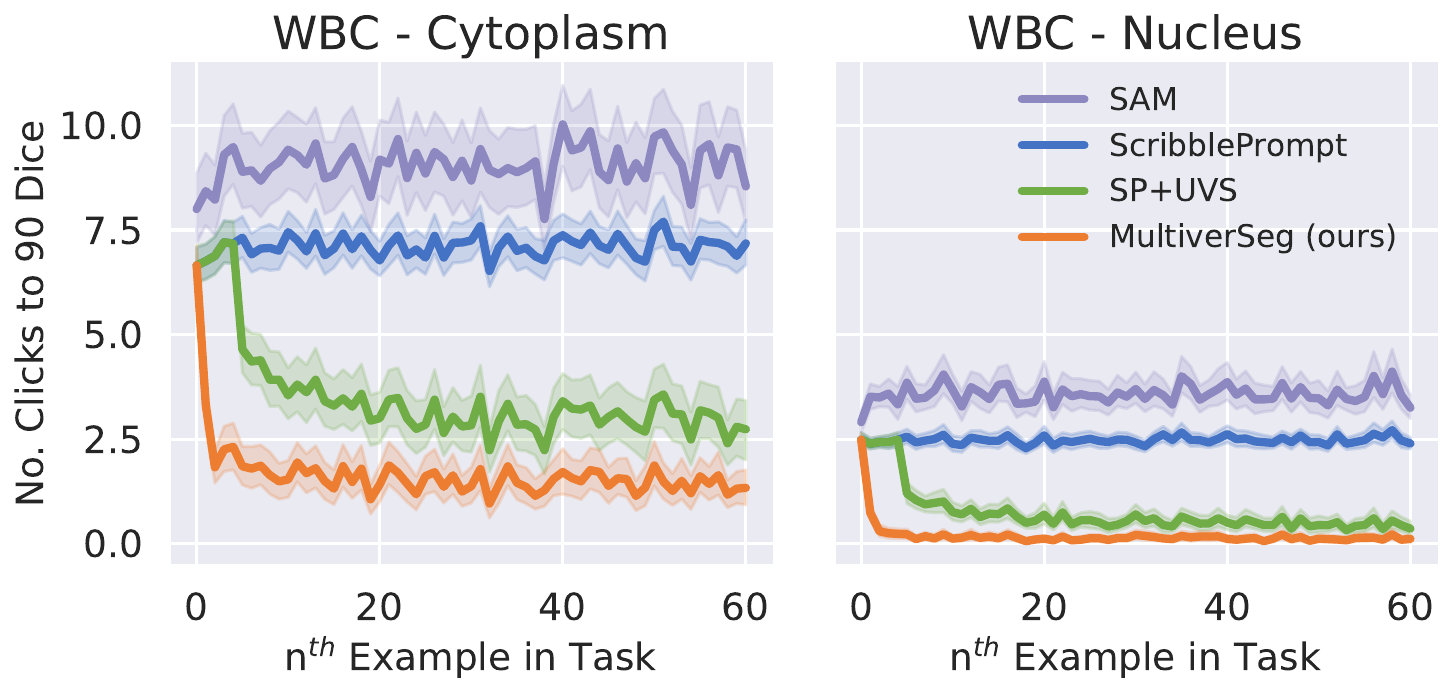}
    \caption{\textbf{Scribble steps to target Dice by task for WBC}. Number of interactions needed to reach a 90\% Dice as a function of the example number being segmented. For the $n^{th}$ image being segmented, the context set has $n$ examples. Shading shows 95\% CI from bootstrapping. WBC~\cite{WBC} is a microscopy dataset containing segmentation tasks for cytoplasm and nuclei of white blood cells. After segmenting a few images from the femur task with \method, the rest of the images in the task can be segmented (to $\geq 90\%$ Dice) with minimal (or no) additional interactions.}
    \label{fig:wbc}
\end{figure}

\begin{figure}
    \centering
    \includegraphics[width=\linewidth]{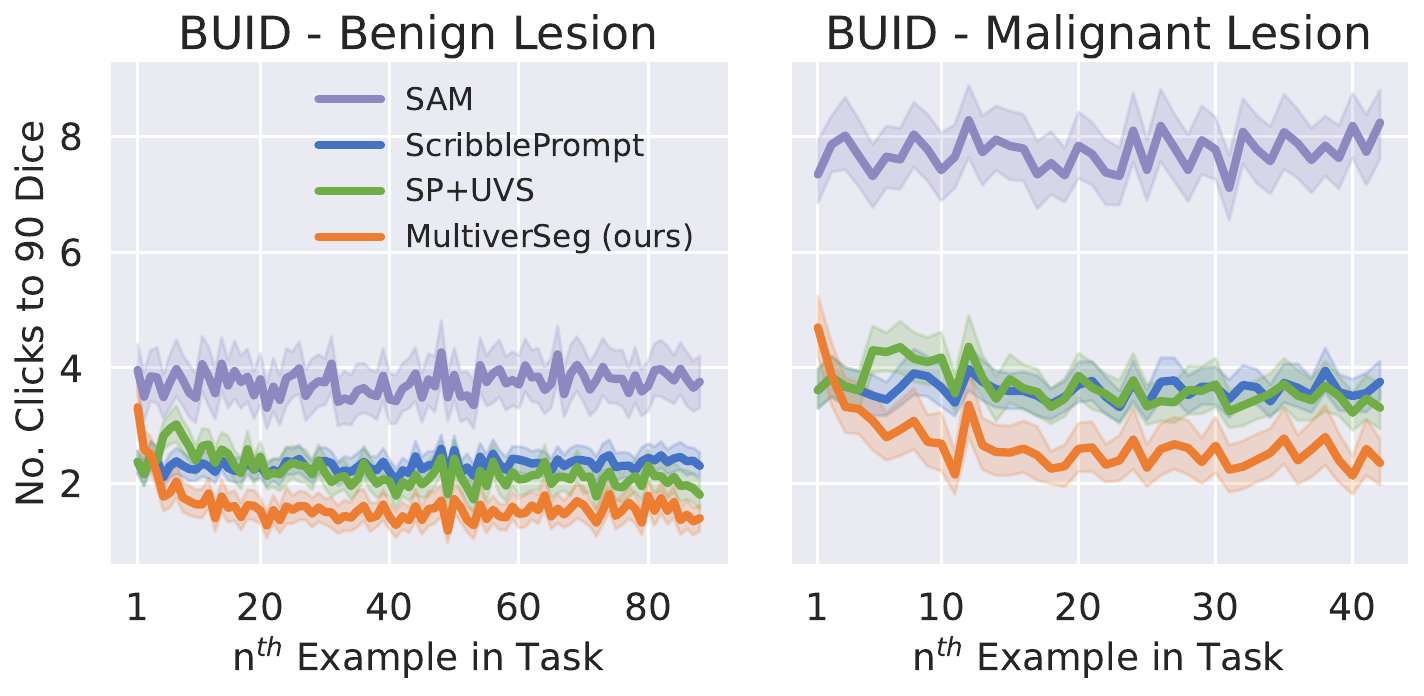}
    \caption{\textbf{Scribble steps to target Dice by task for BUID}. Number of interactions needed to reach a 90\% Dice as a function of the example number being segmented. For the $n^{th}$ image being segmented, the context set has $n$ examples. Shading shows 95\% CI from bootstrapping. BUID~\cite{BUID} is a breast ultrasound dataset containing segmentation tasks for benign and malignant lesions. As the context set of completed segmentations grows, the number of interactions required to segment each additional image with \method gradually declines.}
    \label{fig:buid}
\end{figure}

\begin{figure}
    \centering
    \includegraphics[width=\linewidth]{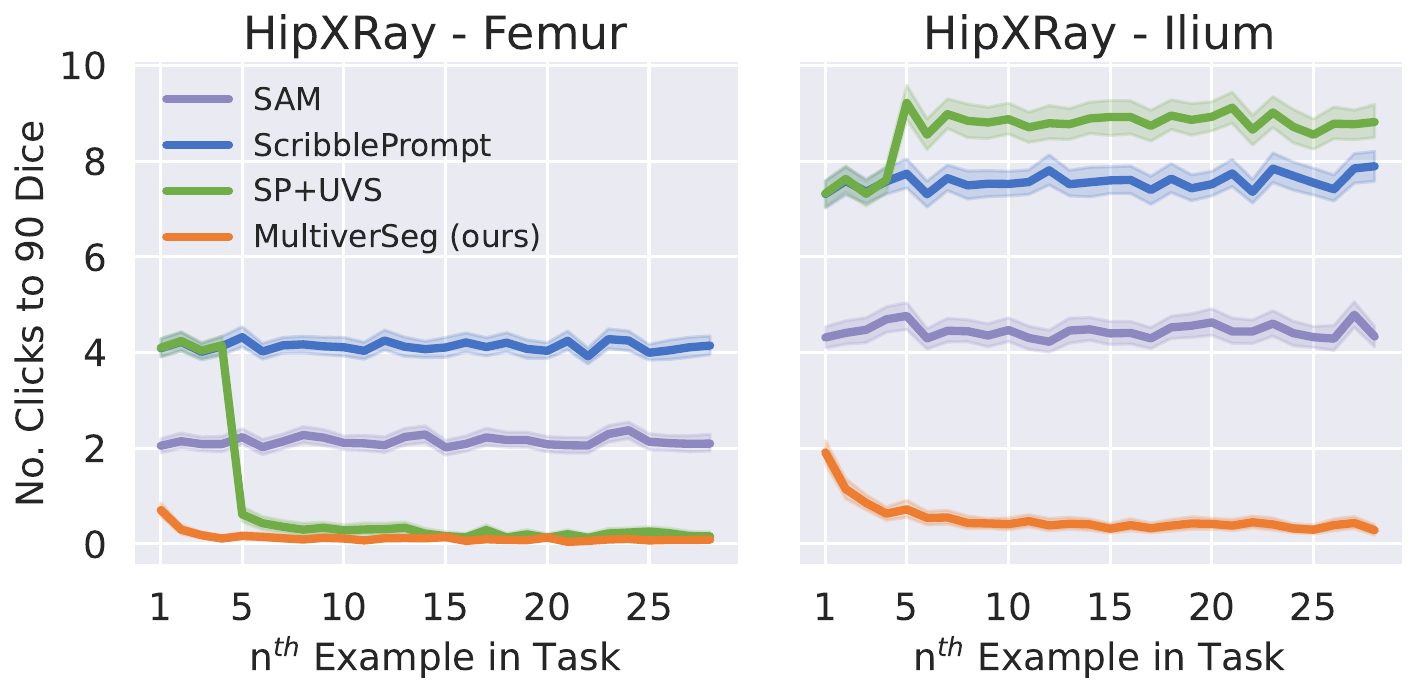}
    \caption{
    \textbf{Center clicks to target Dice by task for HipXRay}. Number of interactions needed to reach 90\% Dice as a function of the example number being segmented. For the $n^{th}$ image being segmented, the context set has $n$ examples. Shading shows 95\% CI from bootstrapping. HipXRay~\cite{HipXRay} is an X-Ray dataset with segmentation tasks for the femur and ilium bones. After segmenting a few images from the femur task with \method, the rest of the images in the task can be segmented (to $\geq 90\%$ Dice) with minimal additional interactions.
    }
    \label{fig:hipxray}
\end{figure}

\subpara{Tasks with more examples}
We show results by task for three datasets with more than 18 test examples per task (\cref{fig:wbc}, \cref{fig:buid}, and \cref{fig:hipxray}). For larger sets of images, using \method results in even greater reductions in the total and average number of user interactions. 

\begin{figure*}
    \centering
    \includegraphics[width=\linewidth]{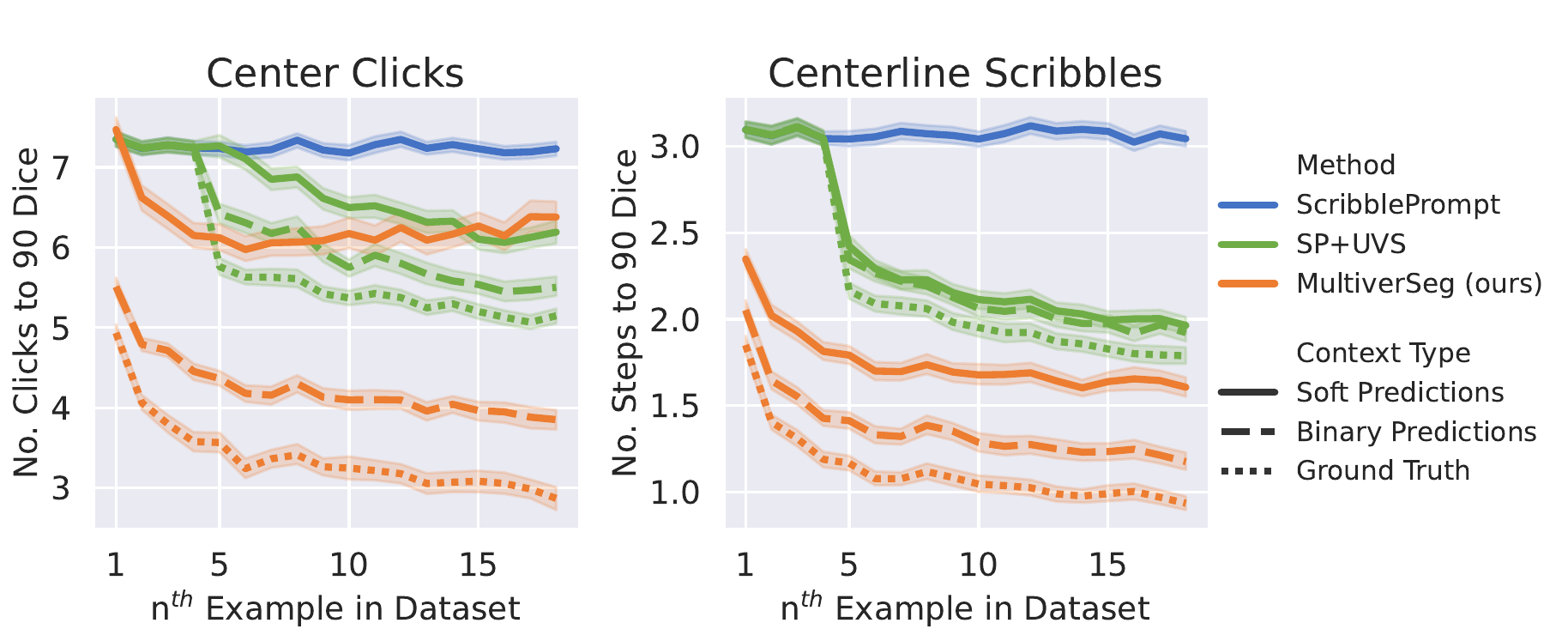}
    \caption{
    \textbf{Interactions to target dice on unseen datasets with different types of context sets}. Number of interactions needed to reach a 90\% Dice as a function of the example number being segmented. For the $n^{th}$ image being segmented, the context set has $n$ examples. We show results with and without thresholding the predictions (``Binary Predictions'' vs. ``Soft Predictions'') . We expect the number of interactions with ``Ground Truth'' context to be a lower bound on the number of interactions to reach 90\% Dice. We show results averaged across validation tasks.
    }
    \label{fig:val_dataset_per_data_variations}
\end{figure*}

\subpara{Context Set Quality}
For \method and SP+UVS, thresholding the predictions before adding them to the context set improved performance (\cref{fig:val_dataset_per_data_variations}). We use the validation split of our validation data (at $128^2$ resolution) to select the best approach (soft or binary predictions in the context set) for each method. 

\method does not perform well when the context set contains \emph{soft} predictions from previous examples, likely because it was trained with ground truth context labels. The number of interactions to 90\% Dice is lowest when the context set contains ground truth labels, however this is not realistic in practice. 

\begin{figure*}
    \centering
    \includegraphics[width=\linewidth]{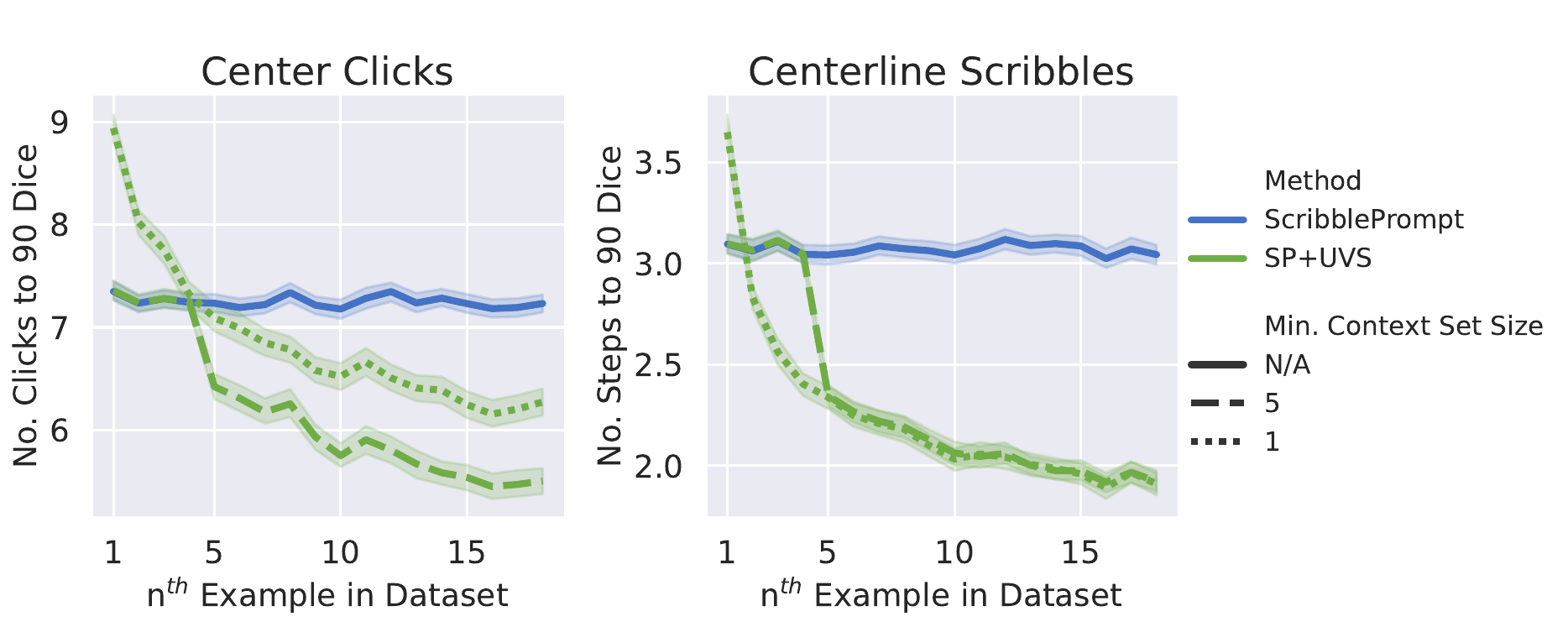}
    \caption{\textbf{Variations of SP+UVS}. Number of interactions needed to reach a 90\% Dice as a function of the example number being segmented. For the $n^{th}$ image being segmented, the context set has $n$ examples. We show results for SP+UVS with different minimum context set size cutoffs, along with ScribblePrompt for reference. SP+UVS with a minimum context set size of $k$, means that when the context set has fewer than $k$ examples, we perform interactive segmentation with ScribblePrompt (ignoring the context examples). When the context set is larger than the minimum size, we first make an in-context segmentation prediction using UniverSeg and then correct that prediction with ScribblePrompt. 
    For small context set sizes, UniverSeg does not make accurate predictions, and initializing ScribblePrompt with UniverSeg's prediction increases the number of interactions required to reach 90\% Dice. We show results averaged across validation tasks.}
    \label{fig:sp_uvs_variations}
\end{figure*}

\subpara{SP+UVS}
Consistent with the original published results, we find that UniverSeg has poor performance for small context sets and initializing ScribblePrompt using the UniverSeg prediction hurts performance when the context set is small. In our final evaluation of SP+UVS, we set the minimum context set size to be 5 examples:
when the context sets contains fewer than 5 examples, we ignore the context and only use ScribblePrompt to make predictions. \cref{fig:sp_uvs_variations} shows variations of SP+UVS with different minimum context set sizes on validation data at $128^2$ resolution. 

\subpara{Total Interactions}
\cref{tab:overall_dice_tot_interactions} shows the total number of interactions, average Dice score, and average 95th percentile Hausdorff distance across all tasks.

\begin{figure*}
    \centering
    \rowcolors{2}{white}{gray!15}
    \begin{tabular}{llccc}
    \toprule
    Interaction Protocol & Method & Dice Score $\uparrow$ & HD95 $\downarrow$ &   Total Steps $\downarrow$ \\
    \midrule
    Center Clicks & SAM-Med2D &  $85.88 \pm 0.14$ &   $3.76 \pm 0.22$ &  $215.58 \pm 2.22$ \\
                     & IMIS-Net &  $81.38 \pm 0.30$ &  $13.05 \pm 0.79$ &  $255.47 \pm 2.53$ \\
                     & SAM &  $90.40 \pm 0.06$ &   $1.40 \pm 0.03$ &  $152.55 \pm 1.76$ \\
                     & SegNext &  $90.50 \pm 0.05$ &   $1.84 \pm 0.06$ &  $158.16 \pm 0.95$ \\
                     & ScribblePrompt &  $90.80 \pm 0.08$ &   $1.48 \pm 0.04$ &  $137.10 \pm 1.21$ \\
                     & SP+UVS &  $90.70 \pm 0.09$ &   $1.49 \pm 0.06$ &  $122.01 \pm 1.93$ \\
                     & MultiverSeg (ours) &  $\mathbf{91.40 \pm 0.14}$ &   $\mathbf{1.26 \pm 0.11}$ &   $\mathbf{87.18 \pm 1.92}$ \\
                     \midrule
    Centerline Scribbles & SAM-Med2D &  $29.58 \pm 3.92$ &  $26.42 \pm 3.36$ &  $178.00 \pm 1.19$ \\
                     & IMIS-Net &  $80.93 \pm 0.40$ &   $3.43 \pm 0.32$ &  $123.46 \pm 2.85$ \\
                     & SAM &  $80.19 \pm 0.74$ &  $19.79 \pm 1.78$ &  $125.14 \pm 2.56$ \\
                     & ScribblePrompt &  $88.19 \pm 0.24$ &   $\mathbf{1.44 \pm 0.06}$ &  $100.70 \pm 2.67$ \\
                     & SP+UVS &  $88.57 \pm 0.23$ &   $1.44 \pm 0.07$ &   $92.50 \pm 1.95$ \\
                     & MultiverSeg (ours) &  $\mathbf{88.65 \pm 0.22}$ &   $1.49 \pm 0.13$ &   $\mathbf{75.23 \pm 1.50}$ \\
    \bottomrule
    \end{tabular}
    \caption{
    \textbf{Average segmentation quality and total interactions per unseen task}. We measure average segmentation quality across a set of 18 test images using Dice score and 95th percentile Hausdorff distance (HD95). For each metric, we show mean and standard deviation from bootstrapping. Dice and HD95 are similar across methods because we simulate interactions until the predicted segmentation has $\geq90\%$ Dice or the maximum number of interaction steps is reached. \method requires the fewest interaction steps per task on average. We report results on images at $256^2$ resolution from 200 simulations.
    }
    \label{tab:overall_dice_tot_interactions}
\end{figure*}

\subsection{Bootstrapping In-Context Segmentation}
\label{appendix:bootstrap_uvs}

\subpara{Setup}
For UniverSeg~\cite{universeg}, a non-interactive in-context segmentation method, we segment the dataset by bootstrapping from a single context example with ground truth segmentation. For each image in the dataset, we make an in-context prediction and then add the prediction to the context set for the next image until all images in the dataset have been segmented. As an upper bound on performance, we also evaluated using ground truth labels in the context set instead of previously predicted segmentations (``UniverSeg (oracle)'').

\begin{figure*}[ht]
    \centering
    % Subfigure for the image
    \begin{subfigure}[b]{0.45\linewidth}
        \centering
        \includegraphics[width=\linewidth]{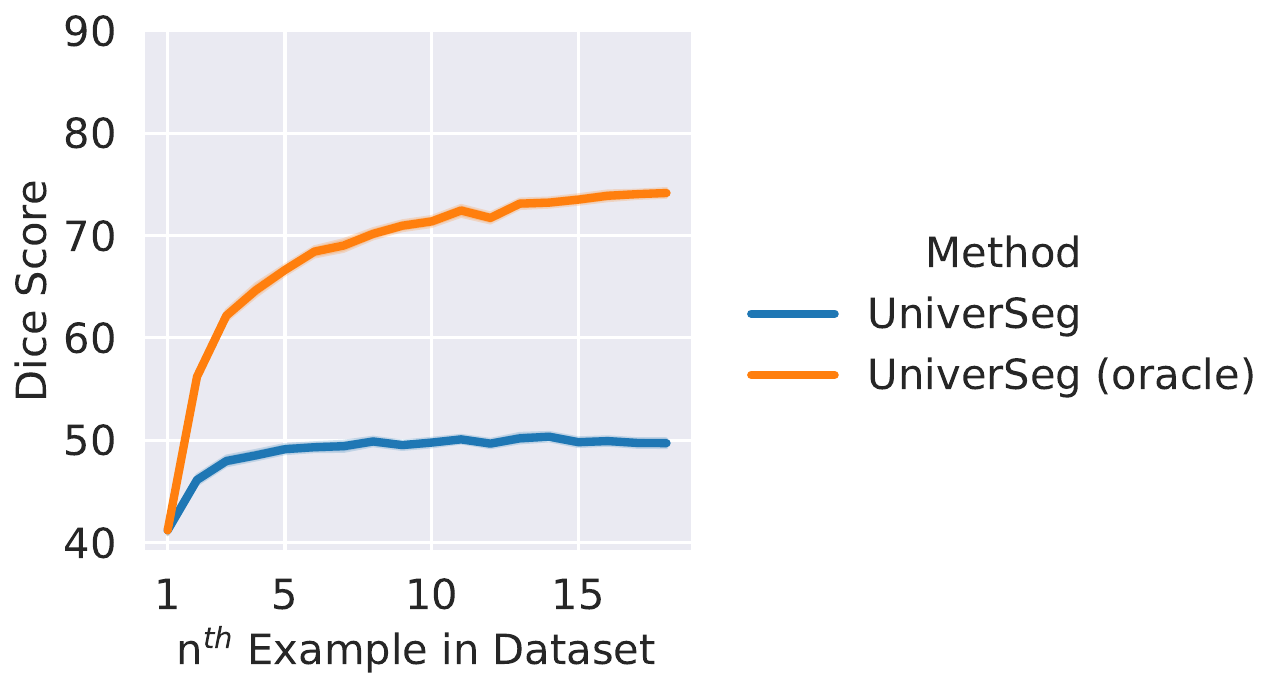}
        \caption{\textbf{Dice score by example number}. We show average Dice Score across unseen test data by example number. We exclude the initial seed example, such that for the $n^{th}$ image being segmented, the context set has $n$ examples.}
        \label{fig:bootstrapping_uvs}
    \end{subfigure}%
    \hfill
    % Subfigure for the table
    \begin{subfigure}[b]{0.5\linewidth}
        \centering
        \rowcolors{2}{white}{gray!15}
        \begin{tabular}{lll}
        \toprule
        Method & Dice Score $\uparrow$ & No. Failures $\downarrow$ \\
        \midrule
        UniverSeg & $48.89 \pm 1.87$ &             $16.76 \pm 0.40$ \\
        UniverSeg (oracle) & $68.15 \pm 1.00$ &             $13.58 \pm 0.24$ \\
        \bottomrule
        \end{tabular}
        \vspace{1em}
        \caption{\textbf{Average performance on unseen tasks}. We report average Dice score per task of 18 images and the average number of examples where the Dice score was less than 90\%.  We report standard deviation across 200 simulations.
        \vspace{\baselineskip}
        }
        \label{tab:uvs_number_failures}
    \end{subfigure}
    \caption{\textbf{Bootstrapping UniverSeg}. 
    We use UniverSeg to sequentially segment images starting from a single example with a ground truth segmentation. After segmenting each image, the image and predicted segmentation are added to the context set for the next example. For the ``oracle'' version, we use ground truth labels in the context set instead of previously predicted segmentations. Even when using ground truth labels in the context set, which we expect to be an upper bound on performance, it was not possible to achieve 90\% Dice for most images.
    }
\end{figure*}

\begin{figure*}
    \centering
    \includegraphics[width=0.9\linewidth]{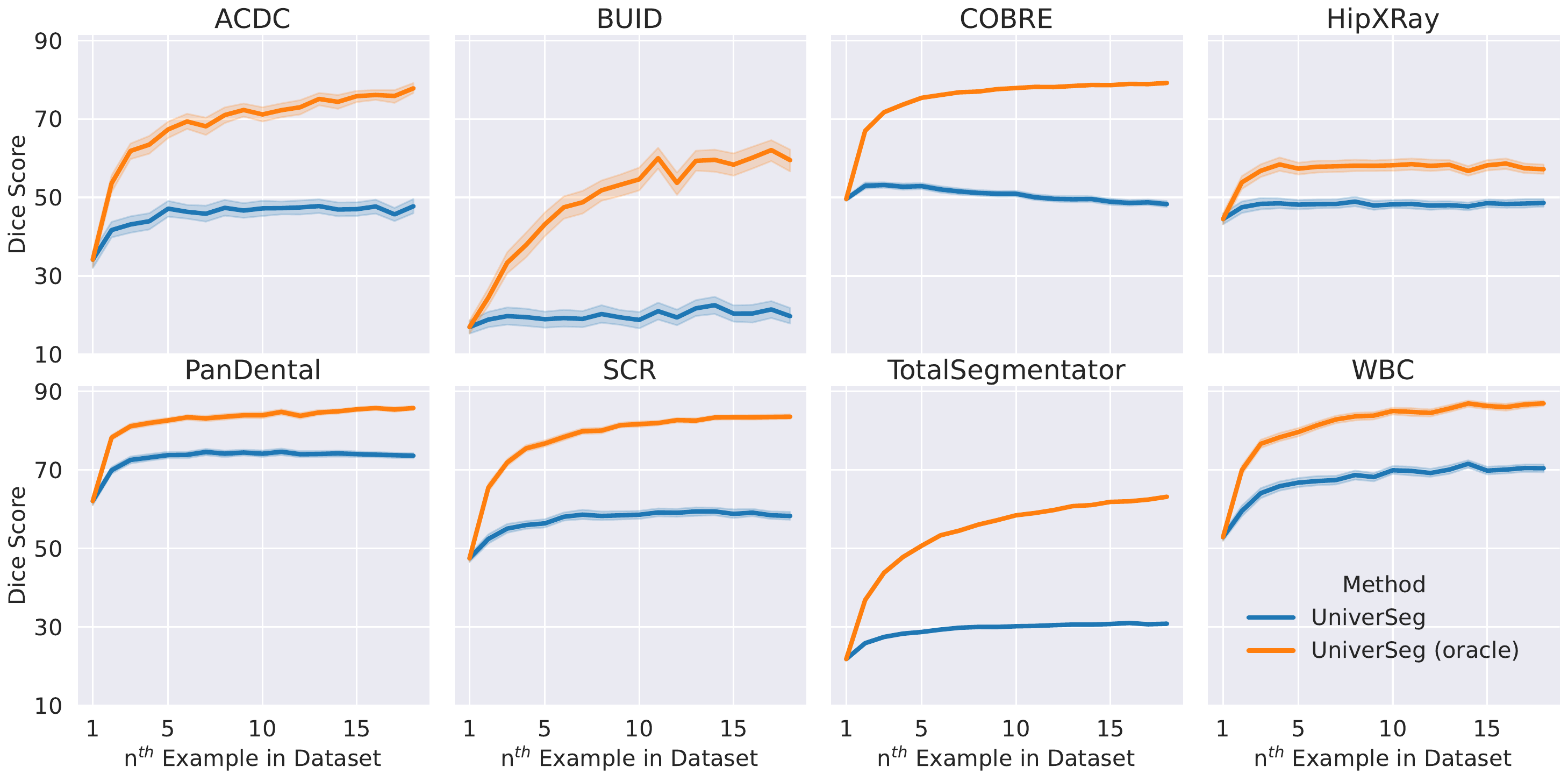}
    \caption{\textbf{Bootstrapping UniverSeg results by dataset}.
    We show Dice score vs. example number for unseen tasks averaged by dataset. After segmenting each image, the image and predicted segmentation are added to the context set for the next example. For the ``oracle'' version, we use ground truth labels in the context set instead of previously predicted segmentations. We exclude the initial seed example, such that for the $n^{th}$ image being segmented, the context set has $n$ examples. Shaded regions show 95\% CI from bootstrapping.
    }
    \label{fig:bootstrapping_uvs_by_dataset}
\end{figure*}

\subpara{Results}
This approach did not produce accurate results, likely because UniverSeg has poor performance for small context sets and/or context sets with imperfect labels (\cref{fig:bootstrapping_uvs}). Because UniverSeg does not have a mechanism to incorporate corrections, it was not possible to achieve 90\% Dice for most images (\cref{tab:uvs_number_failures}). \cref{fig:bootstrapping_uvs_by_dataset} shows results by individual dataset.

\subpara{Context Set Quality}
As with other methods (\method and SP+UVS), we experimented with thresholding the predictions at 0.5 before adding them to the context set. For UniverSeg, thresholding the predictions did not improve Dice scores compared to using the soft predictions in the context set.

%---------------------------------------------------------
\subsection{Comparison to Few-Shot Fine-Tuning}
\label{appendix:finetuning}
%---------------------------------------------------------

 One approach to segmenting a new dataset is to (interactively) segment a few images using a pre-trained foundation model, and then use those examples to train a task-specific interactive segmentation model by fine-tuning the foundation model. In this experiment, we simulated this process using ScribblePrompt. 

\subpara{Setup}
For each task and random seed, we sampled 5 random test examples, and used ScribblePrompt to segment those images using simulated random center clicks. For each image of the 5 images, random center clicks were used to prompt ScribblePrompt until a maximum of 20 clicks was reached or the prediction surpassed 90\% Dice. Then we used those newly labeled images to fine-tune ScribblePrompt from pre-trained weights. We randomly split the 5 images into 4 training examples and 1 validation example.

We fine-tuned ScribblePrompt using the same training interaction protocol, loss function, and data augmentations (\cref{appendix:data_aug}) as \method minus synthetic task augmentations. Each task-specific model was fine-tuned for 300 epochs using the Adam optimizer with a learning rate of $1e^{-6}$ and batch size of $4$. These hyperparameters were selected based on experiments with learning rate $\in \{1e^{-4}, 1e^{-5}, 1e^{-6}\}$ and batch size $\in \{4,8\}$ using the cytoplasm segmentation task from the WBC~\cite{WBC} dataset. For each training run the best checkpoint was selected based on the validation example and then used to interactively segment 13 more test images (to complete the set of 18). 

We repreated this procedure of labelling images and training tasks-specific models for 5 random seeds for each task. Due to the large number of tasks-specific models trained for this experiment, we trained and evaluated on images at $128^2$ to reduce training time.

\subpara{Runtime}
Fine-tuning ScribblePrompt to produce \emph{each} task-specific interactive segmentation model took on average 20 minutes on a NVIDIA A100 GPU. 
In contrast, \method's inference time is $<150$ milliseconds, even with a context set size of 64 examples (\cref{appendix:runtime}). 

\begin{figure*}
    \centering
    \includegraphics[width=\linewidth]{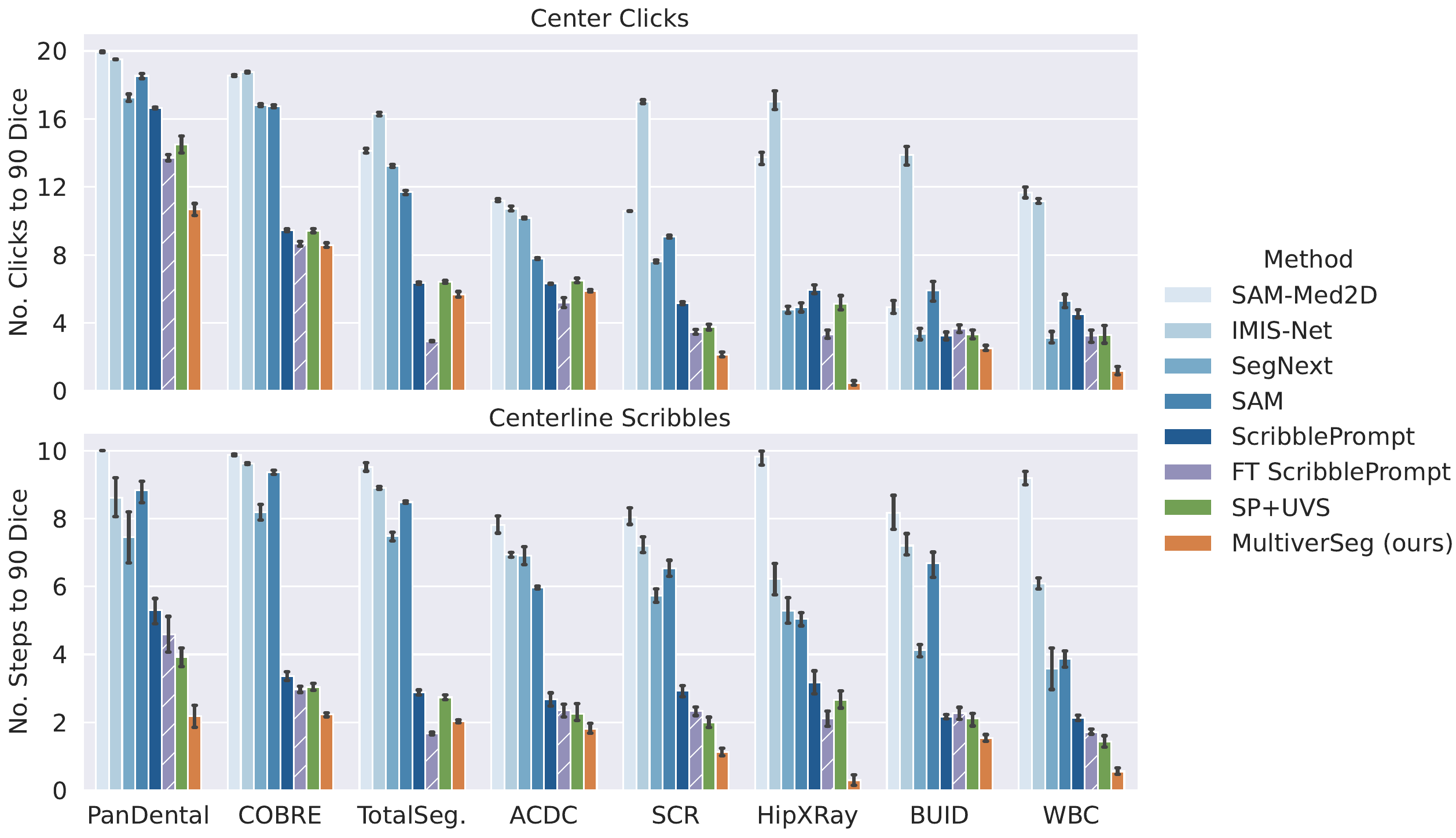}
    \caption{\textbf{\method outperforms task-specific fine-tuning on most datasets}. We show average number of clicks and scribble steps per image to segment 18 images to $\geq90\%$ Dice for each method. For \emph{FT ScribblePrompt} (shaded), we used ScribblePrompt to interactively segment 5 images and then used those examples to fine-tune ScribblePrompt before interactively segmenting the rest. MultiverSeg required fewer interactions thant fine-tuned ScribblePrompt in 13 out of 14 scenarios. Error bars show 95\% CI accross 5 random seeds.}
    \label{fig:finetune}
\end{figure*}

\subpara{Results}
\cref{fig:finetune} shows MultiverSeg required fewer interactions than fine-tuning ScribblePrompt in 13 out of 16 scenarios. On average, the fine-tuning approach required $5.90 \pm 0.10$ clicks or $2.63 \pm 0.13$ scribble steps per image. \method required fewer interactions: $4.64 \pm 0.10$ clicks or $4.64 \pm 0.10$ scribble steps per image.
 
%-
\subsection{Resolution Sensitivity Analysis}
\label{appendix:res128}

We conduct a sensitivity analysis, evaluating \method and the baseline methods at $128^2$ resolution. 

\begin{figure*}
    \centering
    \includegraphics[width=\linewidth]{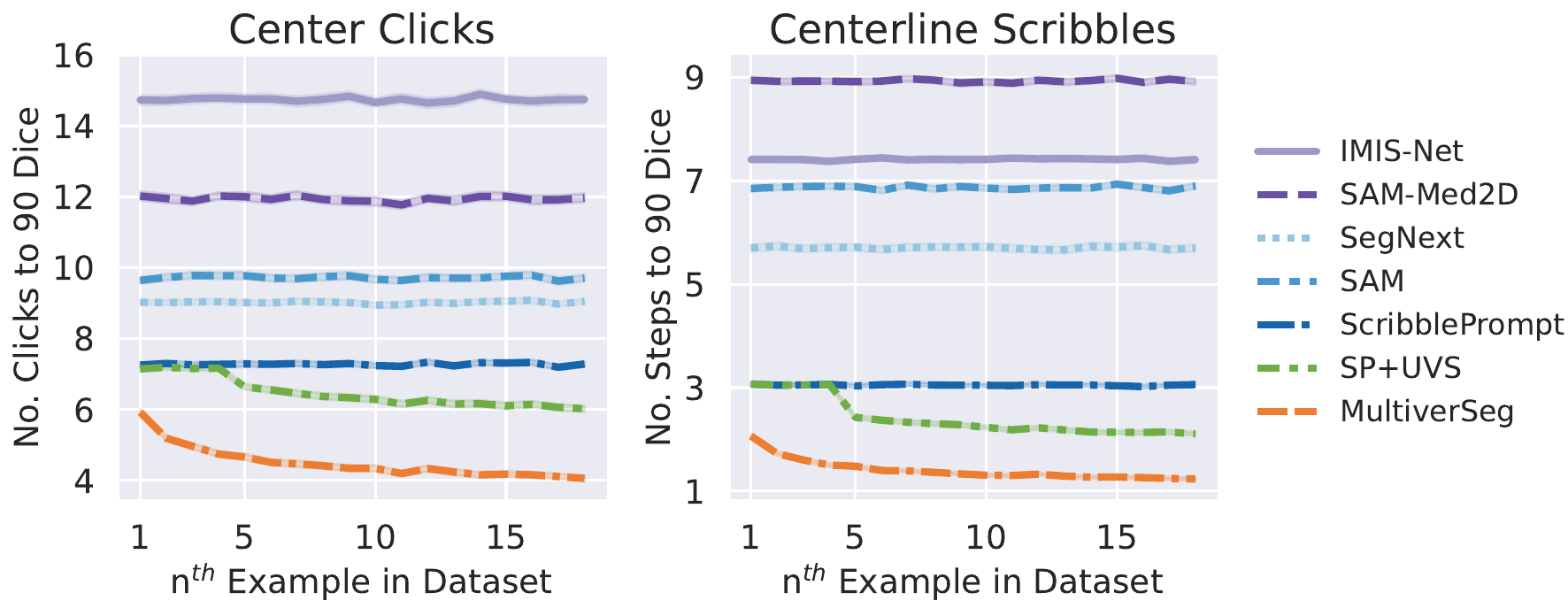}
    \caption{\textbf{Interactions to target Dice on unseen tasks at $\mathbf{128^2}$ resolution}. Number of interactions needed to reach a 90\% Dice as a function of the example number being segmented. For the $n^{th}$ image being segmented, the context set has $n$ examples. \method requires substantially fewer number of interactions to achieve 90\% Dice than the baselines, and as more images are segmented, the average number of interactions required decreases dramatically. Shaded regions show 95\% CI accross 200 random seeds. 
    }
    \label{fig:dataset_interactions_vs_subject-128}
\end{figure*}

\subpara{Results}
\method outperforms the baselines with greater margins when evaluated at $128^2$ resolution compared to $256^2$ resolution. As more examples are segmented and the context set grows, the number of interactions required to get to $90\%$ Dice (NoI90) on the $\text{n}^\text{th}$ example using \method decreases substantially (\cref{fig:dataset_interactions_vs_subject-128}). 

\begin{figure*}
    \centering
    \includegraphics[width=\linewidth]{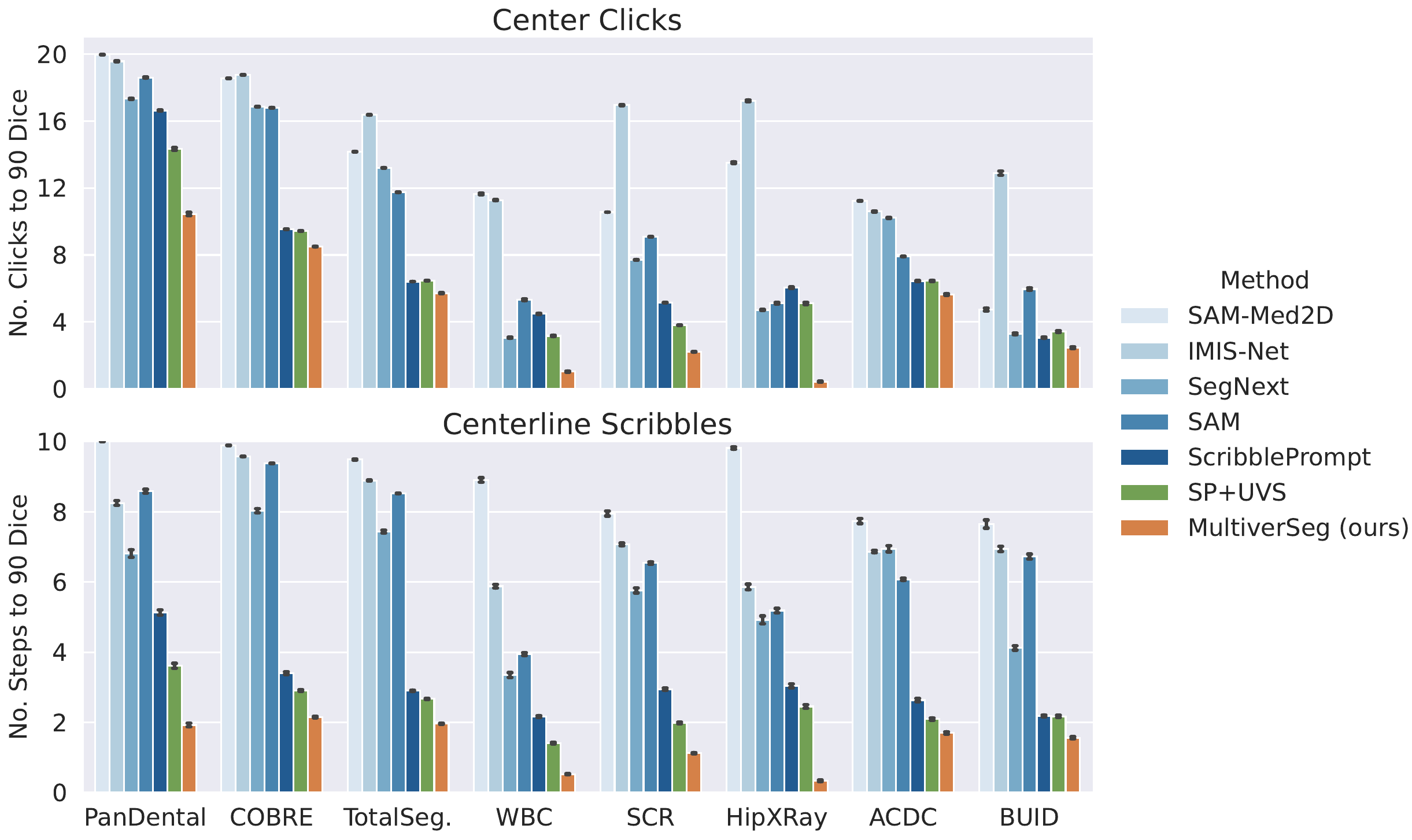}
    \caption{\textbf{Interactions per image by unseen dataset at $\mathbf{128^2}$ resolution}. We show average number of clicks and scribble steps per image to segment 18 images to $\geq90\%$ Dice for each method. In all scenarios, \method required fewer or the same number of interactions than the best baseline. Error bars show 95\% CI accross 200 random seed.}
    \label{fig:total-interactions-128}
\end{figure*}

\method required the fewest number of interactions per image on all datasets (\cref{fig:total-interactions-128}). 
On average, using \method reduced the number of clicks required to segment each dataset by $(36.93 \pm 1.53)\%$ and the number of scribble steps required by $(36.93 \pm 1.53)\%$ compared to ScribblePrompt.

%--------------------------------------------------------
\section{Experiment 2: Analysis}
%--------------------------------------------------------

%--------------------------------------------------------
\subsection{In-Context Segmentation}
%--------------------------------------------------------

\begin{figure*}
    \centering
    \includegraphics[width=\linewidth]{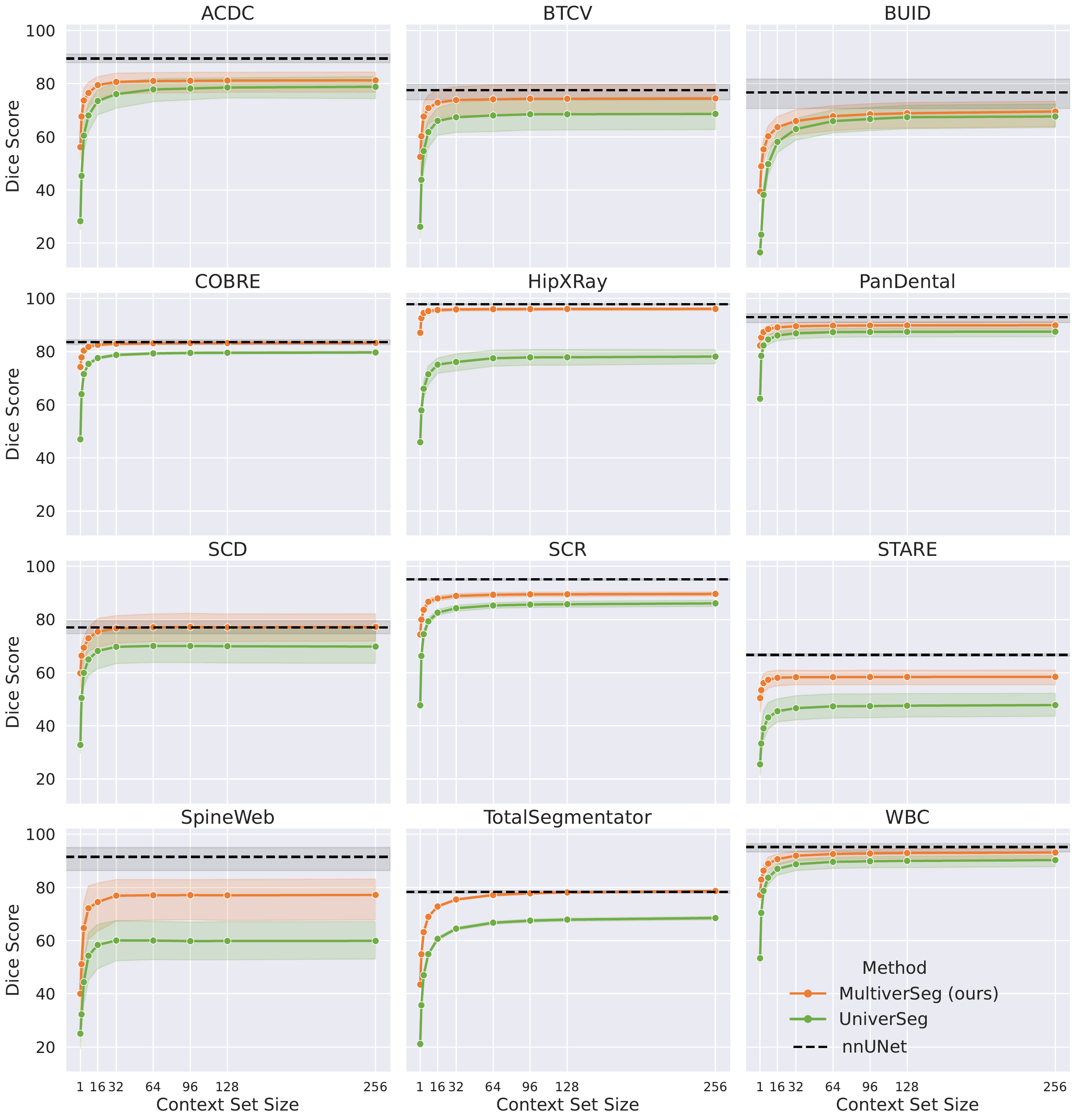}
    \caption{
    \textbf{In-context segmentation performance across context set sizes on unseen datasets}. We compare \method to UniverSeg, an in-context segmentation method, given ground truth context labels. Points show results for context set sizes 1, 2, 4, 8, 16, 32, 64, 96, 128 and 256. Shading shows 95\% CI from bootstrapping.
    }
    \label{fig:uvs_eval_dataset}
\end{figure*}

\subpara{Results} 
\cref{fig:uvs_eval_dataset} show results by dataset with different context set sizes.

%--------------------------------------------------------
\subsection{Interactive Segmentation In Context}
%--------------------------------------------------------

\begin{figure*}
    \centering
    \includegraphics[width=0.8\linewidth]{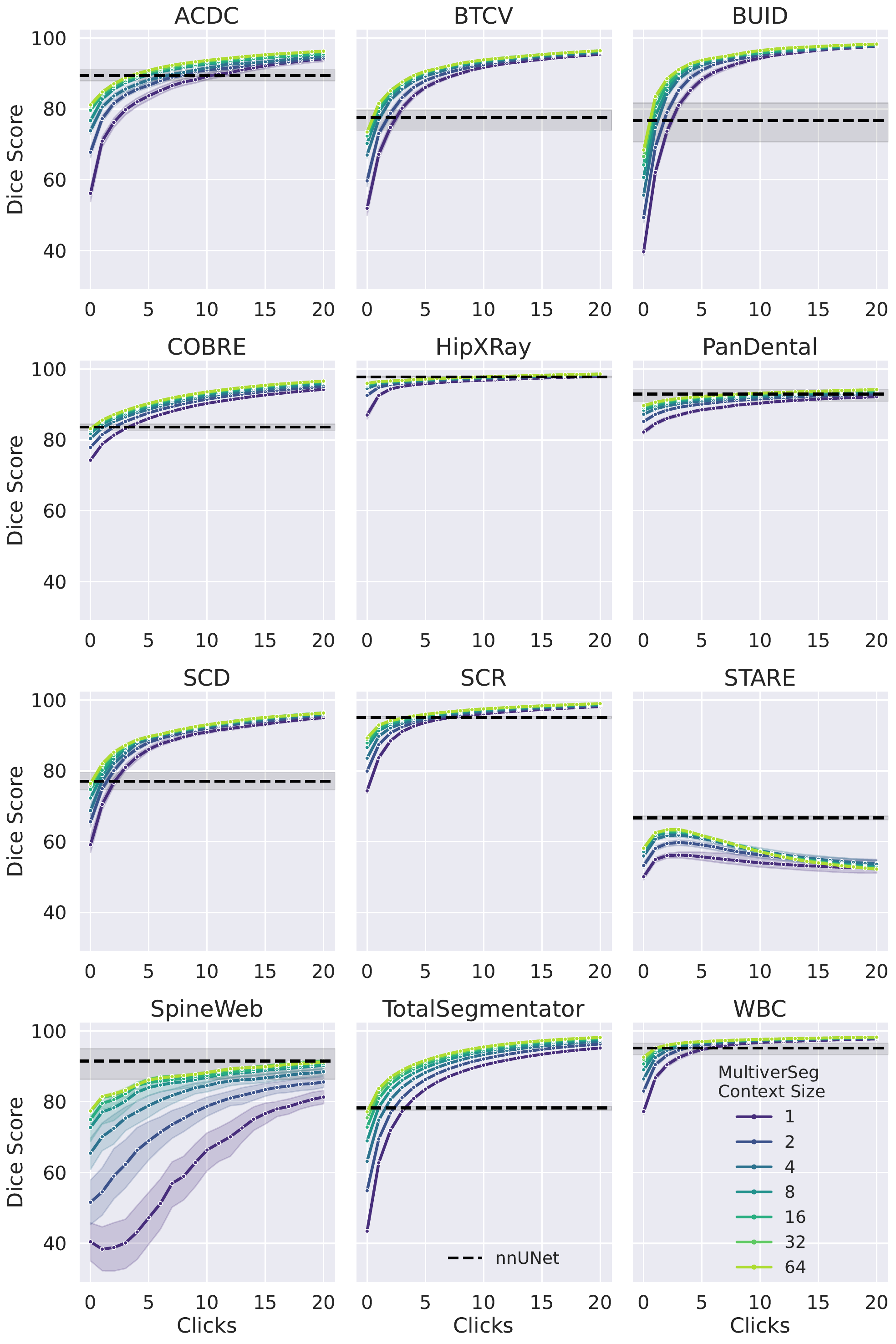}
    \caption{
    \textbf{Interactive segmentation in context with center clicks on unseen datasets}.
    \method's interactive segmentation performance with the same number of interactions improves as the context set size grows. We first make an initial prediction based on the context set (step 0), and then simulate corrections with one center click at a time. Shading shows 95\% CI from bootstrapping.
    }
    \label{fig:mvs_eval_dataset_clicks}
\end{figure*}

\begin{figure*}
    \centering
    \includegraphics[width=0.8\linewidth]{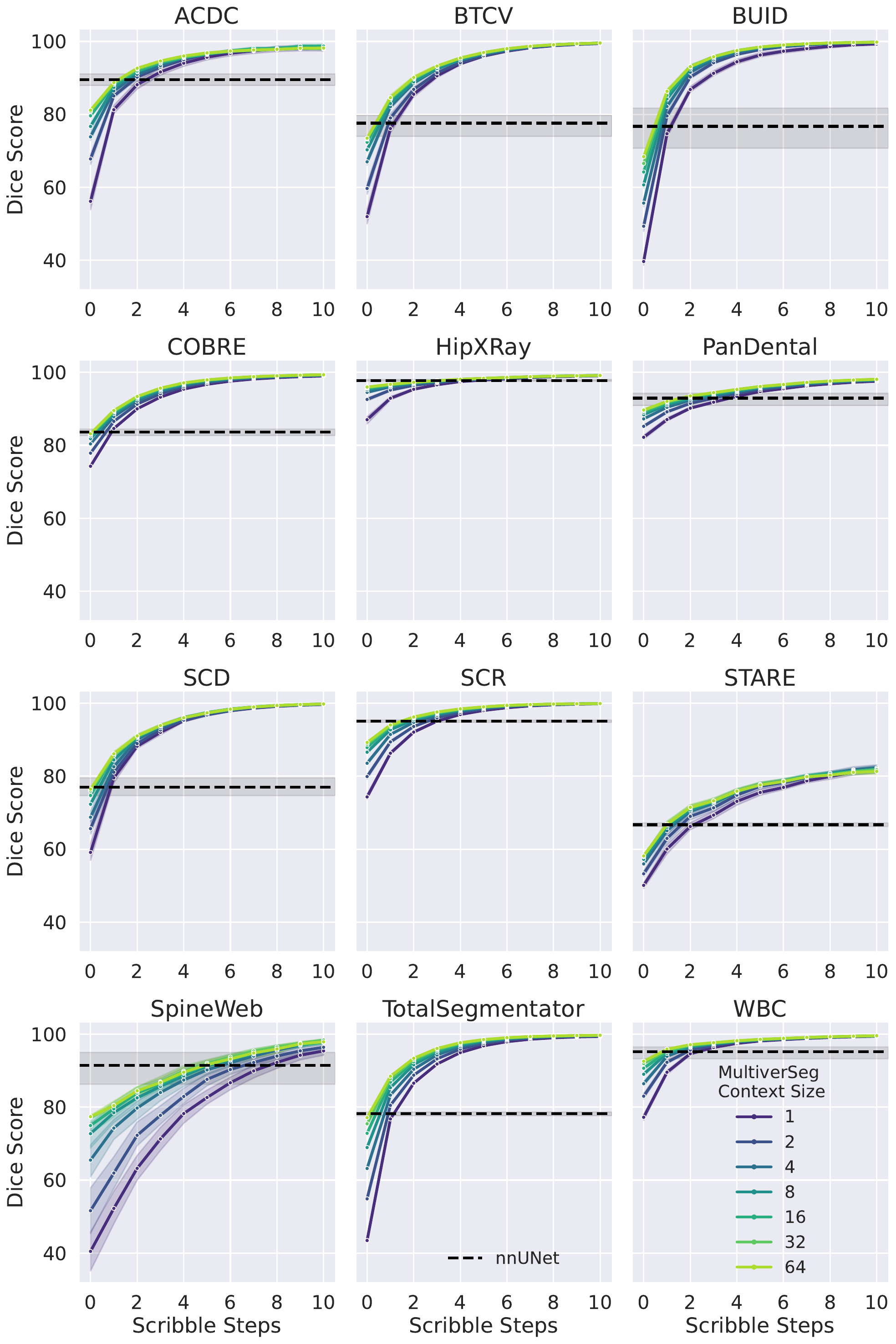}
    \caption{
    \textbf{Interactive segmentation in context with centerline scribbles on unseen datasets}.
    \method's interactive segmentation performance with the same number of interactions improves as the context set size grows. We first make an initial prediction based on the context set (step 0), and then simulate centerline scribble corrections. Shading shows 95\% CI from bootstrapping.
    }
    \label{fig:mvs_eval_dataset_scribbles}
\end{figure*}

\subpara{Results}
\cref{fig:mvs_eval_dataset_clicks} and \cref{fig:mvs_eval_dataset_scribbles} show results by dataset using center clicks and centerline scribbles, respectively. 

%--------------------------------------------------------
\subsection{Inference Runtime and Memory Usage}\label{appendix:runtime}
%--------------------------------------------------------

MultiverSeg's inference runtime scales linearly with the context set size (\cref{tab:runtime}).  However, even with a context set of 64 examples, the runtime is under $150$ms. Prior work on interactive interfaces indicates $<500$ms latency is sufficient for cognitive tasks~\cite{liu2014effects}. Since the interactions are stored in masks, inference runtime (per prediction) is not affected by the number of user interaction inputs.

\begin{table}[!h]
    \centering
    \begin{tabular}{c|c|c}
        \toprule
        Context Size & Inference Time (ms) & GPU Memory \\
        \midrule
        1   & $25.28 \pm 0.16$  & 28 MB \\
        16  & $57.05 \pm 0.20$  & 1.89 GB \\
        32  & $86.57 \pm 0.06$  & 3.64 GB\\
        64  & $146.04 \pm 0.16$ & 7.15 GB \\
        128 & $267.42 \pm 0.24$ & 12.16 GB \\
        256 & $604.15 \pm 0.36$ & 24.17 GB \\
        \bottomrule
    \end{tabular}
    \caption{
    \textbf{Inference runtime and GPU memory usage with different context set (CS) sizes}. We report mean $\pm$ standard deviation runtime in milliseconds across 1,000 predictions at $128^2$ resolution with 1 click on an NVIDIA A100 GPU. GPU memory usage is reported as peak allocated memory during inference.
    }
    \label{tab:runtime}
\end{table}

\end{document}